\documentclass[Afour,sageh,times]{sagej}

\usepackage{bm}
\usepackage{moreverb,url}
\usepackage[colorlinks,bookmarksopen,bookmarksnumbered,citecolor=red,urlcolor=red]{hyperref}

\newcommand\BibTeX{{\rmfamily B\kern-.05em \textsc{i\kern-.025em b}\kern-.08em
		T\kern-.1667em\lower.7ex\hbox{E}\kern-.125emX}}

\usepackage{tikz}
\newcommand{\V}[1]{\boldsymbol{#1}}
\newcommand{\M}[1]{\mathbf{#1}}

\def\volumeyear{2016}

\begin{document}
	
	\runninghead{The Authors}
	
	\title{Versatile, Robust, and Explosive Locomotion with Rigid and Articulated Compliant Quadrupeds}
	
	\author{Jiatao Ding\affilnum{1,2}, Peiyu Yang\affilnum{1}, Fabio Boekel\affilnum{1}, Jens Kober\affilnum{1}, Wei Pan\affilnum{3}, Matteo Saveriano\affilnum{2}, and Cosimo Della Santina\affilnum{1,4}}
	
	\affiliation{\affilnum{1}Cognitive Robotics, Delft University of Technology, The Netherlands\\
		\affilnum{2}Department of Industrial Engineering, University of Trento, Trento, Italy\\
		\affilnum{3}Department of Computer Science, The University of Manchester, UK\\
		\affilnum{4}Institute of Robotics and Mechatronics, German Aerospace Center (DLR), Germany}
	
	\corrauth{Jiatao Ding, and Wei Pan.}
	
	\email{jiatao.ding@unitn.it, wei.pan@manchester.ac.uk}
	
	\begin{abstract}
		Achieving versatile and explosive motion with robustness against dynamic uncertainties is a challenging task. Introducing parallel compliance in quadrupedal design is deemed to enhance locomotion performance, which, however, makes the control task even harder. This work aims to address this challenge by proposing a general template model and establishing an efficient motion planning and control pipeline. To start, we propose a reduced-order template model—the dual-legged actuated spring-loaded inverted pendulum with trunk rotation—which explicitly models parallel compliance by decoupling spring effects from active motor actuation. With this template model, versatile acrobatic motions, such as pronking, froggy jumping, and hop-turn, are generated by a dual-layer trajectory optimization, where the singularity-free body rotation representation is taken into consideration. Integrated with a linear singularity-free tracking controller, enhanced quadrupedal locomotion is achieved. Comparisons with the existing template model reveal the improved accuracy and generalization of our model. Hardware experiments with a rigid quadruped and a newly designed compliant quadruped demonstrate that \textit{i)} the template model enables generating versatile dynamic motion; \textit{ii)} parallel elasticity enhances explosive motion. For example, the maximal pronking distance, hop-turn yaw angle, and froggy jumping distance increase at least by 25$\%$, 15$\%$ and 25$\%$, respectively; \textit{iii)} parallel elasticity improves the robustness against dynamic uncertainties, including modelling errors and external disturbances. For example, the allowable support surface height variation increases by 100$\%$ for robust froggy jumping.
	\end{abstract}
	
	
	\maketitle
	
	\section{Introduction}\label{introduction}
	Quadrupedal robots show a huge possibility of achieving high mobility (\cite{bjelonic2022offline,margolis2024rapid,jenelten2024dtc}). 
	Articulated compliant-quadrupeds (also called elastically actuated quadrupeds, soft quadrupeds) (\cite{della2021soft}) are gaining popularity  (\cite{seidel2020using,hutter2016anymal,bjelonic2023learning,grimmer2012comparison,ding2024roboust}), among which robots with parallel elastic actuators (PEAs) are becoming prominent due to their improved performance, such as providing energetic benefits (\cite{bjelonic2023learning}), reducing peak power (\cite{grimmer2012comparison}), and achieving explosive motion (\cite{ding2024roboust}).
	
	With the quadrupedal maneuver, achieving \textit{versatile}, \textit{robust}, and \textit{explosive} locomotion has been a long-lasting goal:
	\begin{itemize}
		\item {\textit{Versatile locomotion}: performing multi-modal motions with different contact sequences.}
		
		\item \textit{Robust locomotion}: accomplishing desired locomotion tasks albeit with dynamic uncertainties, such as modelling errors and external disturbances.
		
		\item \textit{Explosive locomotion}: realize highly dynamic motion with a flight phase, requiring the robot to output large force/torque in a short period.
	\end{itemize}
	However, achieving this goal is a challenging task because of the complex dynamics and intensive contact switch. Furthermore, the benefits of parallel compliance towards this goal have not been well investigated.
	
	\begin{figure}
		\centering
		\includegraphics[width=\columnwidth-0mm]{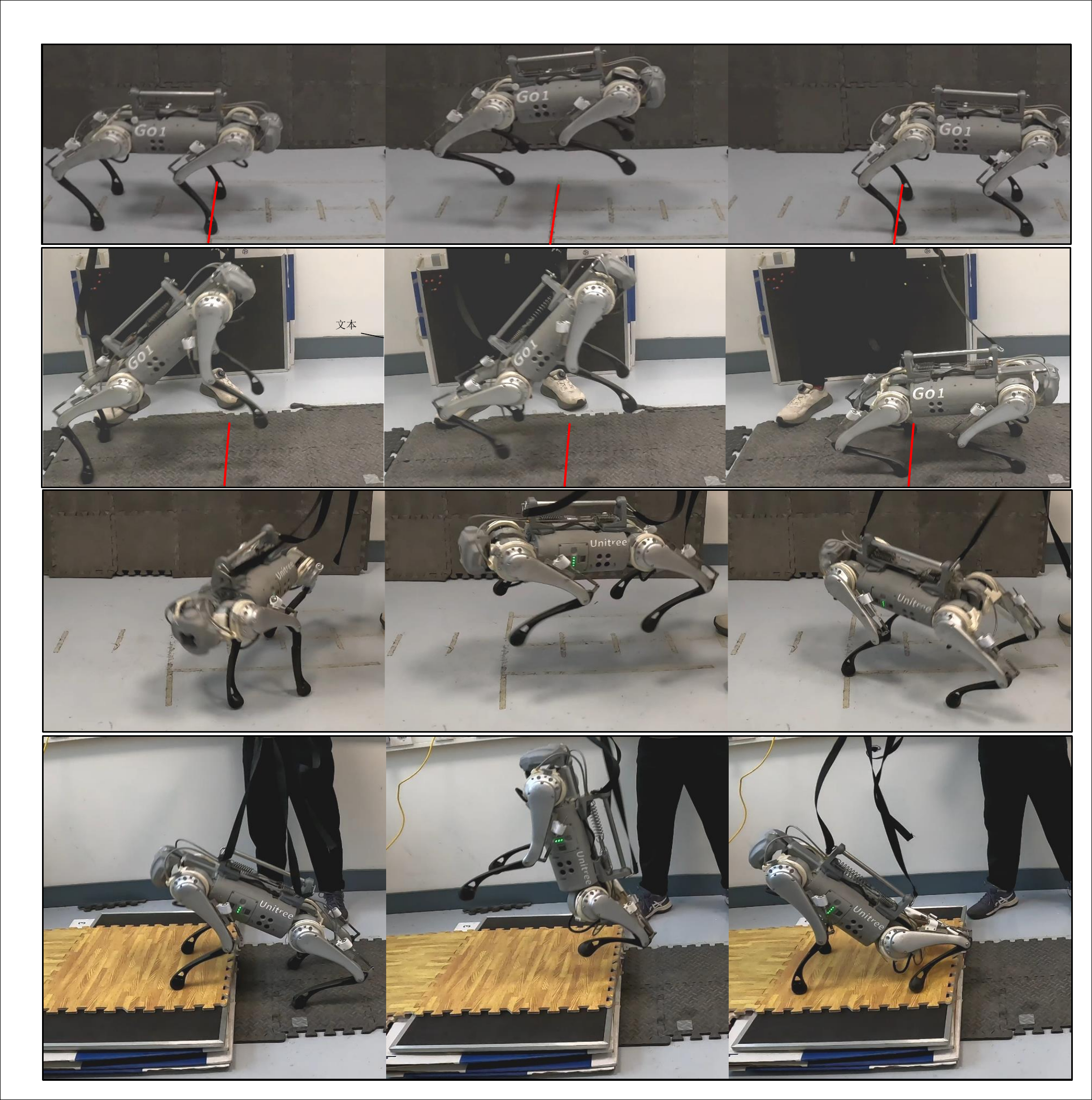}
		\caption{The compliant Go1 with parallel springs (we call it E-Go-V2) performs versatile and explosive motions. From top to bottom, the robot performs \textit{1)} $40\,$cm forward pronking, \textit{2)} $40\,$cm forward froggy jumping, \textit{3)} $145^{\circ}$ clockwise hop-turn, and \textit{4)} robust froggy jumping from the uneven surface. At the bottom, the robot jumps from the non-coplanar surface ($10\,$cm height variation) without knowing the terrain information in advance. The red lines in the first two rows mark the desired landing positions for the rear legs.}
		\label{fig:snaptshots}
	\end{figure}
	
	One of the limitations is the lack of an accurate and general template model which is capable of capturing the locomotion dynamics driven by PEA. 
	As a well-known simplification, the spring-loaded inverted pendulum (SLIP) model captures locomotion dynamics with few open parameters (\cite{2006Compliant}). As a variant, the actuated SLIP (aSLIP) contributes to high-speed walking (\cite{liu2016terrain}), hopping (\cite{xiong2018bipedal}), and running (\cite{green2020planning}).
	However, the above SLIPs, together with many other variants, only identify compliance at the system level without explicitly characterizing parallel elasticity. To address this issue, the work in (\cite{ding2024roboust}) proposed a novel aSLIP model that decouples motor actuation with parallel elasticity. Then, a dual-leg version was proposed in (\cite{ding2024Quadrupedal}), implementing quadrupedal jumping with asymmetric legs. Nevertheless, due to the absence of trunk motion, the SLIP model above is restrictive when performing more acrobatic motions, such as froggy jumping and hop-turn, where a large body rotation is required.

	For motion generation and control, a singularity-free formulation needs to be considered, especially when performing explosive motion with large body rotation (\cite{li2024cafe,yue2024online,nguyen2022continuous}). Although various trajectory optimization (TO) schemes (\cite{posa2014direct}) have been developed to plan acrobatic motion, most of them face the singularity issue due to the utilization of the Euler angle to represent body angles. Some of the work, such as (\cite{nguyen2022contact}), uses the rotation matrix to avoid the Gimbal lock. However, they usually require a well-designed reference trajectory defined in advance and rely on a complex full-body dynamics model. For motion control, although singularity-free representation has been proposed in recent work, such as those adopting model predictive control (MPC)  (\cite{zhang2024robots,garcia2021time,ding2021representation}), only a very limited locomotion modality has been validated. Furthermore, none of the above work achieves dynamic locomotion with articulated compliant quadrupeds.
	
	In this work, we propose \textit{TD-aSLIP}, a novel template model that captures both leg movements and trunk rotation. By decoupling passive compliance from active motor actuation, parallel compliance introduced by mechanical design is explicitly modelled. In particular, the nonlinear parallel compliance with configuration variation is captured, achieving higher accuracy. Then, we develop a dual-layer TO that consists of one coarse SLIP-based optimization and one refined kinodynamics optimization to generate optimal explosive motions, whereby singularity-free rotational movement obeying both kinematics and dynamics constraints is achieved. We demonstrate improved accuracy compared with existing SLIP models and generalization to versatile motions such as pronking, hop-turn, and froggy jumping. Through extensive simulation and hardware experiments\footnote{Results can be seen: 
		{\url{https://youtu.be/6n8YmNtXcp4}}.}, this work also demonstrates improved performance, including enhanced tracking performance and robustness against disturbances by exploiting parallel elasticity.
	
	To the best of our knowledge, this work is among the first to realize versatile, robust, and explosive locomotion with a PEA-driven quadruped. Typical motions are demonstrated in Fig.~\ref{fig:snaptshots}. To summarize, our main contributions are as follows.
	\begin{enumerate}
		\item {A general template model, i.e., TD-aSLIP, is proposed, with the capability of capturing the varying angular momentum and nonlinear parallel compliance.}
		
		\item {A singularity-free motion planning and control scheme is introduced to achieve dynamic motion with large body rotation, which can be applied to both the rigid and compliant cases. In particular, no specific reference is needed as a prerequisite when achieving highly dynamic motions.}
		
		\item {Versatile, robust, and explosive motion is realized with both rigid and articulated compliant quadrupeds. In particular, hardware experiments demonstrate that parallel compliance contributes to enhanced explosive motion with higher robustness.}
		
	\end{enumerate}
	
	The remainder of this paper is organized as follows. 
	Section~\hyperlink{sec_relate_work}{II} details the related work.
	Section~\hyperlink{sec_overview}{III} gives a first glance of the proposed approach. Sections~\hyperlink{sec_slip_model}{IV}, Section~\hyperlink{sec_motion_planner}{V} and Section~\hyperlink{sec_compliance_control}{VI} separately introduce the TD-aSLIP dynamics, dual-layer motion planner, and whole-body control scheme. Section~\hyperlink{sec_evaluation}{VII} and Section~\hyperlink{sec_hardware_evaluation}{VIII} separately present simulation and hardware experiments. Section~\hyperlink{sec_conclusion}{IX} concludes this work and discusses the future direction.
	
	\section{Related Work}
	\label{sec_relate_work}
	
	\subsection{Quadrupedal locomotion with template model}\label{mode_qua_jump}
	Dynamic quadrupedal locomotion can be realized via model-based control (\cite{ma2020bipedal,bjelonic2022offline,zhao2024survey,wensing2023optimization}). Based on full-body dynamics, explosive motions have been achieved via TO, as done in (\cite{nguyen2019optimized,gilroy2021autonomous,nguyen2022continuous}). To reduce the computational burden, reduced-order template models, such as the single rigid body (SRB) model (\cite{chignoli2021online,ding2021representation,gu2024high}) and the centroidal dynamics model (\cite{jeon2022online,chignoli2022rapid,ding2020kinodynamic}), are used.  Although effective, the above models did not capture the passive compliance introduced by mechanical design.
	
	In contrast, the biology-inspired SLIP model consists of only a body center and a leg with a linear spring in it (\cite{1989The}). Using very few open parameters, it captures key dynamics properties while characterizing the system compliance. However, the assumption of constant spring stiffness and a fixed rest leg length in canonical 2D and 3D SLIPs (see \cite{2006Compliant,hong2024slip,wensing2013high,han20223d}) limits its application. To address this issue, researchers introduced the ``actuated'' SLIP (aSLIP), allowing the change of the rest length (\cite{rezazadeh2015spring,liu2015dynamic,wang2022fast}), the spring constant (\cite{2018Coupling}), or the force rules (\cite{green2020planning}). Although widely adopted in legged locomotion (see \cite{lakatos2018dynamic,calzolari2022single,xiong20223}), the lack of trunk motion limits the application to quadrupedal locomotion requiring large body rotations. To enrich the expressiveness of the model, trunk SLIPs (TSLIPs)~(\cite{sharbafi2013robust,drama2020trunk,ding2023robust}) explicitly consider the torque acting on the hip joint. However, the presented SLIPs fail to explicitly capture the parallel compliance introduced by mechanical design. To exploit parallel compliance, a novel aSLIP model was proposed in (\cite{ding2024roboust}), where the system actuation and parallel compliance are decoupled. Then, in (\cite{ding2024Quadrupedal}), a dual-leg version was introduced. However, the aSLIP models in (\cite{ding2024roboust}) and (\cite{ding2024Quadrupedal}) assume a fixed spring constant, resulting in a coarse reflection of mechanical compliance. In addition, the trunk motion is not considered.
	
	In this work, we introduce {TD-aSLIP}, a general aSLIP model with double legs and a rotational trunk for versatile and dynamic locomotion. As a result, the change of angular momentum is characterized. Inherited from the previous work (\cite{ding2024Quadrupedal}), this enhanced TD-aSLIP model also exploits parallel compliance by decoupling it with the system actuation on each leg. Furthermore, a configuration-aware compliance mapping method is proposed to better exploit parallel elasticity. 
	
	\begin{figure*}
		\centering
		\includegraphics[width=2\columnwidth-10mm]{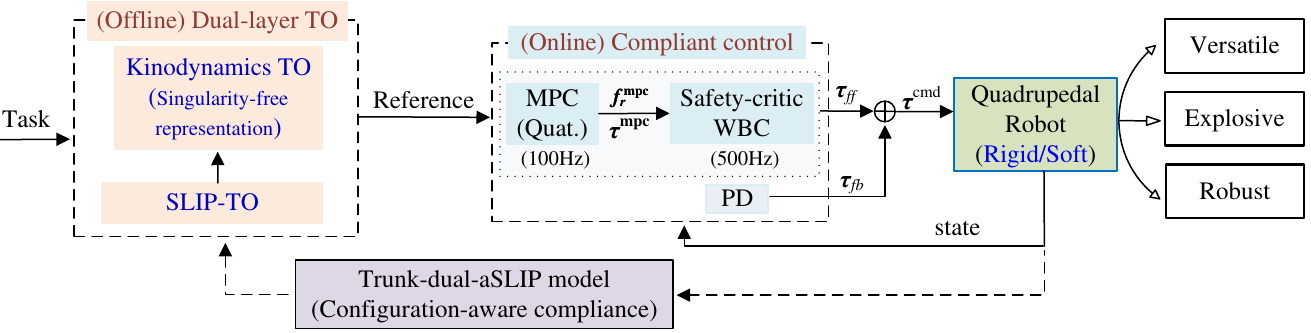}
		\caption{{Optimization-base motion planning and control for versatile, robust, and explosive quadrupedal locomotion. The singularity-free formulation is incorporated into trajectory optimization and compliant control.}
		}
		\label{fig:control_algorithm}
		\vspace{-5mm}
	\end{figure*} 
	
	\subsection{Explosive motion with large body rotation}
	TO-based strategies are widely applied to generate dynamic motion in quadrupeds (\cite{gilroy2021autonomous,jeon2022online,chignoli2022rapid,ding2020kinodynamic,chignoli2021online,gu2024high}), among which the motion with large body rotation, such as rotation jumping and froggy jumping, has gained a lot of attention recently. To achieve this kind of motion, it is crucial to capture the angular dynamics  (\cite{nguyen2019optimized,nguyen2022continuous,papatheodorou2024momentum,zhou2022momentum,nguyen2022contact}). In order to obtain optimal trajectory, the above optimization approaches require a well-defined reference trajectory (in Cartesian or joint space). Also, most of the above work assumes a fixed contact timing. To alleviate this, the work in (\cite{song2022optimal}) utilized an evolutionary algorithm to search the trajectory, which, however, relied on a complex prioritized fitness function. On the other hand, the above methods utilized Euler angles to represent the body movement, which may face the Gimbal lock when performing motions with large orientation variations.
	To avoid this, the work in (\cite{nguyen2022contact}) utilized the rotation matrix to represent the body movement, where, however, a hand-tuned reference is again required.
	
	To accomplish desired tasks, model-based tracking control approaches like MPC and whole-body control (WBC), have been widely used to track the reference~(\cite{neunert2018whole,bellicoso2018dynamic,li2024cafe,kim2019highly}). To avoid the singularity caused by large body movements, the work in (\cite{ding2021representation}) used the rotation matrix, while the work in (\cite{zhang2024robots,garcia2021time}) used the quaternion dynamics to capture the desired body rotation. However, since no compliant dynamics are considered, the above work can not achieve explosive motion with articulated soft quadrupeds with parallel compliance.  
	
	In this work, we propose a unified motion planning and control pipeline for explosive motion with both rigid and articulated soft quadrupeds. By utilizing a dual-layer optimization scheme, versatile motions are generated without requiring well-defined references. Through hierarchical control, explosive motion with robustness is achieved.  In particular, quaternion dynamics are incorporated in formulating both the motion planner and controller, avoiding the singularity caused by the large body rotation.

	\subsection{Quadrupeds with parallel compliance} By introducing passive compliance in mechanical design, articulated compliant quadrupeds, also called soft or elastic quadrupeds, have gained attention recently for achieving efficient and dynamic motion  (\cite{hutter2016anymal,calzolari2023embodying,vezzi2024two,pollayil2022planning,badri2022birdbot}). The elastic actuation, especially PEA, has shown potential in saving energy (\cite{della2020exciting,roozing2021efficient,zhang2024novel}), reducing peak torque (\cite{grimmer2012comparison}), and enhancing explosive motion (\cite{liu2024novel}).
	
	To investigate the benefits of parallel compliance in quadrupedal locomotion, compliant quadrupeds have been proposed. By installing parallel springs in the knee joints, PUPPY II showed an adaptive frequency oscillating motion (\cite{buchli2006finding}). The compliant Cheetah-cub (\cite{sprowitz2013towards}) and Oncilla (\cite{sprowitz2018oncilla}), with springs attached to the knee joints, achieved fast trotting or running trotting. Through the combination of reinforcement learning and Bayesian optimization, the optimal design of parallel springs was realized for the enhanced ANYmal robot (\cite{bjelonic2023learning}), realizing energy-efficient and robust trotting. Furthermore, with parallel springs acting independently on all sagittal joints, Morti (\cite{ruppert2022learning}) learned to perform energy-efficient locomotion. 
	The work in (\cite{ding2024Quadrupedal}) developed a reproducible platform by adding parallel springs on a commercially available quadruped and then validated the dynamic motion with model-based control. Although energy-efficient and dynamic locomotion has been reported in the above approaches, more versatile and explosive motions, such as froggy jumping with large body rotation, have not been investigated. In (\cite{arm2019spacebok}), highly explosive jumping was shown with the PEA-driven SpaceBok, but only in an environment with lower gravity.  
	
	In this work, we highlight parallel compliance for explosive locomotion. We start by proposing a new hardware design, providing an easy-reproducible platform with parallel compliance. Then, through extensive simulation and hardware experiments, we fully validate the benefits of parallel compliance in achieving explosive motion.
	
	\section{Methodology overview}\label{sec_overview}
	
	This section provides a first glance at the proposed approach, as illustrated in Fig.~\ref{fig:control_algorithm}. For brevity, we briefly introduce the motion generation and control pipeline.

	In the motion planning stage, we optimize the multi-modal dynamic motions based on the TD-aSLIP model. Through a dual-layer TO scheme, feasible motions satisfying both the kinematics and dynamics constraints are obtained. In particular, the singularity-free trajectory is generated for the desired task without requiring a hand-tuned reference. Due to the utilization of the novel SLIP model with configuration-aware stiffness, parallel compliance introduced by the mechanical add-ons is explicitly exploited.

	Then, a hybrid torque control strategy including feedback impedance control ($\bm{\tau}_{fb}$) and feedforward torque ($\bm{\tau}_{ff}$) compensation is utilized to generate commanded torque ($\bm{\tau}^{\text{cmd}}$). In particular, a hierarchical structure is used, where a quaternion MPC strategy is first adopted to compute ground reaction forces (GRFs, denoted as $\bm{f}^{\text{mpc}}_r$) and the corresponding torque ($\bm{\tau}^\text{mpc}$), which are then passed to the WBC to generate the joint torques. Differing from the classic WBC, the parallel compliance is explicitly compensated by whole-body dynamics. Safety constraints are also incorporated into the WBC formulation to ensure safe behavior.
	
	\textit{Notations:} In this work, matrices and vectors are noted in bold fonts. The superscript $(\cdot)^{\text{T}}$ represents the transpose operation. For the matrix with multiple rows and columns (noted in the bold normal font), the subscript $(\cdot)_{(k)}$ means the $k$-th column and the subscript $(\cdot)_{(j,k)}$ notes the element at the $j$-th row and $k$-th column. For the vector (with the size $n \times 1$ or $1 \times n$ that is noted in the bold italic font), $(\cdot)_{(k)}$ refers to the $k$-th element. $\M I_n$ is the identity matrix with the size of $n \times n$. Variables accompanied with $(\cdot)^{r}$ and $(\cdot)^{e}$ separately denote the reference and estimated values. {Besides, variables with $\underline{(\cdot)}$ and $\overline{(\cdot)}$ separately denote the upper and lower boundaries.} {Also, we index the four legs using the pair: {\text{front left} $\rightarrow$ \text{FL}: 1}, front right $\rightarrow$ \text{FR}: 2, rear left $\rightarrow$ \text{RL}: 3, rear right $\rightarrow$ \text{RR}: 4.}
	
	\section{TD-aSLIP with stiffness variation}\label{sec_slip_model} 
	In this section, we derive the TD-aSLIP dynamics. In particular, we explicitly characterize the nonlinear parallel stiffness caused by the configuration variation.
	
	\subsection{TD-aSLIP dynamics}
	
	\begin{figure}
		\centering
		\includegraphics[width=\columnwidth]{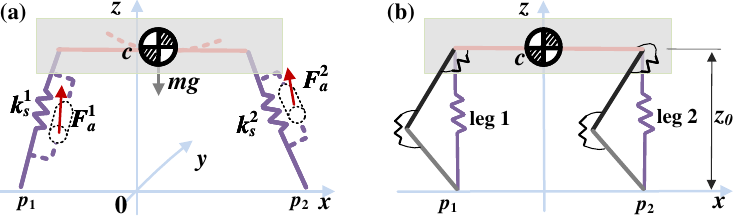}
		\caption{The TD-aSLIP model with parallel elasticity, (a) temple model, and (b) quadrupedal motion mapping in the homing pose.}
		\label{fig:quadrupedal_slip_model}
	\end{figure}
	
	For legged robots, we assume a lumped mass in the trunk center, with massless legs. 
	{Considering trunk rotation, we have the following \textbf{stance dynamics} for the TD-aSLIP model (see Fig.~\ref{fig:quadrupedal_slip_model}(a))}
	\begin{equation} \label{equ:trunk_slip_dual_legs}
		\begin{aligned}
			m\ddot{\bm{c}} &= {\bm{F}^1 + \bm{F}^2}+ m\bm{g},  \\
			\bm{\mathcal{{I}}}\ddot{\bm{\theta}} &\approx {(\bm{p}_1 - \bm{c})\times \bm{F}^1 + (\bm{p}_2 - \bm{c})\times \bm{F}^2},        \end{aligned}  \\
	\end{equation}
	with
	\begin{equation} \label{equ:force_distribution}
		\begin{aligned}
			\bm{F}^1 &= \bm{F}_s^1 + \bm{F}_a^1,  \quad
			\bm{F}^2 = \bm{F}_s^2 + \bm{F}_a^2,        \end{aligned}  \\
	\end{equation}
	where $\bm{F}^1 \in \mathbb{R}^3$ and $\bm{F}^2 \in \mathbb{R}^3$ are the GRFs on leg 1 and leg 2, respectively. $\bm{F}_a^1 \in \mathbb{R}^3$ and $\bm{F}_a^2 \in \mathbb{R}^3$ are the actuation forces in leg 1 and leg 2, respectively. $\bm{F}_s^1 \in \mathbb{R}^3$ and $\bm{F}_s^2 \in \mathbb{R}^3$ are the elastic forces on leg 1 and leg 2, respectively, ${\V{c}} \! \in \! \mathbb{R}^{3}$, ${\V{p}_1} \! \in \! \mathbb{R}^{3}$, and ${\V{p}_2} \! \in \! \mathbb{R}^{3}$ are the center of mass (CoM) position, foot position of leg 1, and foot position of leg 2, $\V{g} \! \in \! \mathbb{R}^{3}$ is the gravitational acceleration, $\ddot{\V{c}} \! \in \! \mathbb{R}^{3}$ and $\ddot{\bm{\theta}} \! \in \! \mathbb{R}^{3}$ separately denote the linear acceleration and angular acceleration of trunk center. $\bm{\mathcal{{I}}} \! \in \! \mathbb{R}^{3\times3}$ is the inertial matrix, $m \in \mathbb{R}$ is the total mass, and $\times$ denotes the cross-product operation. 
	
	In \eqref{equ:trunk_slip_dual_legs}, assuming massless legs, the inertia matrix is computed as:
	\begin{equation}
		\bm{\mathcal{{I}}} = \mathbf{R} \, \bm{\mathcal{{I}}}_b \, \mathbf{R}^{\text{T}},
	\end{equation}
	where $\bm{\mathcal{{I}}}_b$ is the inertia tensor in the body frame and $\mathbf{R}$ is the rotation matrix from the body frame to the global frame.
	
	For the unidirectional spring, the spring forces $\bm{F}_s^1$ and $\bm{F}_s^2$ are determined by
	\begin{equation} \label{equ:assp1_slip2}
		\begin{aligned}
			\bm{F}_s^1 = {k_s^1(\text{max}(l_0^1 -||\bm{l}_1||_2,0))\hat{\bm{{l}}}_1},\\
			\bm{F}_s^2 = {k_s^2(\text{max}(l_0^2 -||\bm{l}_2||_2,0))\hat{\bm{{l}}}_2},
		\end{aligned}
	\end{equation}
	where $k_s^1\in \mathbb{R}$ and $k_s^2\in \mathbb{R}$ separately denote the equivalent spring constant on leg 1 and leg 2,  $\bm{l}_1 = \bm{c} - \bm{p}_1 \in \mathbb{R}^3$ and $\bm{l}_2= \bm{c} - \bm{p}_2 \in \mathbb{R}^3$ are the leg vectors. $\hat{\bm{l}}_1 \in \mathbb{R}^3$  and $\hat{\bm{l}}_2 \in \mathbb{R}^3$ are the unit leg vectors, $l_0^1 \in \mathbb{R}$ and $l_0^2 \in \mathbb{R}$ are the rest lengths, and $\operatorname{||\cdot||_2}$ indicates the L$_2$ norm of a vector.

	\textbf{In flight}, the trunk motion is determined by
	\begin{equation} \label{equ:trunk_slip_dual_legs_flight}
		\begin{aligned}
			\ddot{\bm{c}} = \V{g}, \quad
			\M{I}\ddot{\bm{\theta}} = 0.
		\end{aligned}
	\end{equation}
	
	\subsection{Configuration-aware stiffness mapping}
	
	Since parallel elasticity is decoupled with motor actuation, the above template model can be applied to both rigid (without parallel springs) and articulated compliant (with parallel springs) cases. 
	However, in \eqref{equ:assp1_slip2},  the $k_s^1$ and $k_s^2$ characterize the spring constant in Cartesian space, while the spring constants in hardware are expressed in the joint space.
	Thus, we need to identify the equivalent $k_s^1$ and $k_s^2$ to capture the parallel compliance contributed by the joint movement.
	
	\subsubsection{Equivalent spring constant}
	For a quadruped, assuming a small variation in the joint angles of each leg, the potential energy (${V}_\text{quad}^\text{spring}$) contributed by spring deformation is calculated as
	\begin{equation}
		\setlength{\abovedisplayskip}{3pt}
		\setlength{\belowdisplayskip}{3pt}
		\begin{aligned}
			{{V}_\text{quad}^\text{spring}=  \frac{1}{2} (\delta\V{q}^j)^{\text{T}}\mathbf{k}^j(\delta\V{q}^j),}
		\end{aligned}
		\label{potential_quadrupedal}
	\end{equation}
	where $\delta\V{q}^j\in \mathbb{R}^3$ denotes the joint angle variations of the $j$-th leg ($j \in \{1,...,4\}$), $\mathbf{k}^j \in \mathbb{R}^{3\times3}$ is a diagonal matrix taking the spring constants of the parallel springs installed on the hip, thigh, and calf joints as the diagonal elements.
	
	Given the small variation in the $j$-th leg length {($\delta \bm{l}^j$)}, the potential energy (${V}_\text{slip}^\text{spring}$) contributed by the spring tension is
	\begin{equation}
		\setlength{\abovedisplayskip}{3pt}
		\setlength{\belowdisplayskip}{3pt}
		\begin{aligned}
			{V}_\text{slip}^\text{spring}\!=\! \frac{1}{2}(\delta \bm{l}^j)^{\text{T}} \mathbf{k}^j_\text{equ} (\delta \bm{l^j}) 
			\!=\! {\frac{1}{2}(\delta\bm{q}^j)^{\text{T}}[(\mathbf{J}^j)^{\text{T}} \mathbf{k}_\text{equ}^j \mathbf{J}^j](\delta\bm{q}^j),}
		\end{aligned}
		\label{potential_slip}
	\end{equation}
	where $\mathbf{k}_\text{equ}^j\in \mathbb{R}^{3\times3}$ is the equivalent spring constant matrix and $\mathbf{J}^j\in \mathbb{R}^{3\times3}$ is the Jacobian matrix. 
	
	To match the Cartesian-space stiffness with the joint-space stiffness, we demand ${V}_\text{quad}^\text{spring}={V}_\text{slip}^\text{spring}$. As a result, for the $j$-th leg, we have equivalent stiff matrix ($\mathbf{k}_\text{equ}^j$) computed as
	\begin{equation}
		\setlength{\abovedisplayskip}{3pt}
		\setlength{\belowdisplayskip}{3pt}
		\begin{aligned}
			\mathbf{k}_\text{equ}^j=((\mathbf{J}^j)^{-1})^{\text{T}} \mathbf{k}^j(\mathbf{J}^j)^{-1}.
		\end{aligned}
		\label{potential_slip_match}
	\end{equation}
	
	To map a quadruped to the TD-aSLIP, we first pair the four legs into two groups. Taking the sagittal motion as an example, we can pair the two front legs of the quadruped into leg 2 and the two rear legs into leg 1, as illustrated in Fig.~\ref{fig:quadrupedal_slip_model}(b). Then, we assume that the two legs within in each pair share the same joint variations and the same Jacobian matrix due to symmetry. 
	In addition, the two legs share the same spring parameters. 
	
	As a result, the equivalent spring matrix of the TD-aSLIP model is double the results in \eqref{potential_slip_match}. Therefore, the scalar spring constant ($k_s^i$) of the TD-aSLIP is determined as
	\begin{equation}
		\setlength{\abovedisplayskip}{3pt}
		\setlength{\belowdisplayskip}{3pt}
		\begin{aligned}
			k_s^{i}= 2{\sqrt{{(\mathbf{k}^{i}_\text{equ(0,0)})^2 + (\mathbf{k}^{i}_\text{equ(1,1)})^2+ (\mathbf{k}^{i}_\text{equ(2,2)})^2}},}
		\end{aligned}
		\label{potential_slip_match_element}
	\end{equation}
	with $i \in \{1, 2\}$ marking the leg 1 or 2 of the TD-aSLIP.
	
	\begin{figure}
		\centering
		\includegraphics[width=\columnwidth]{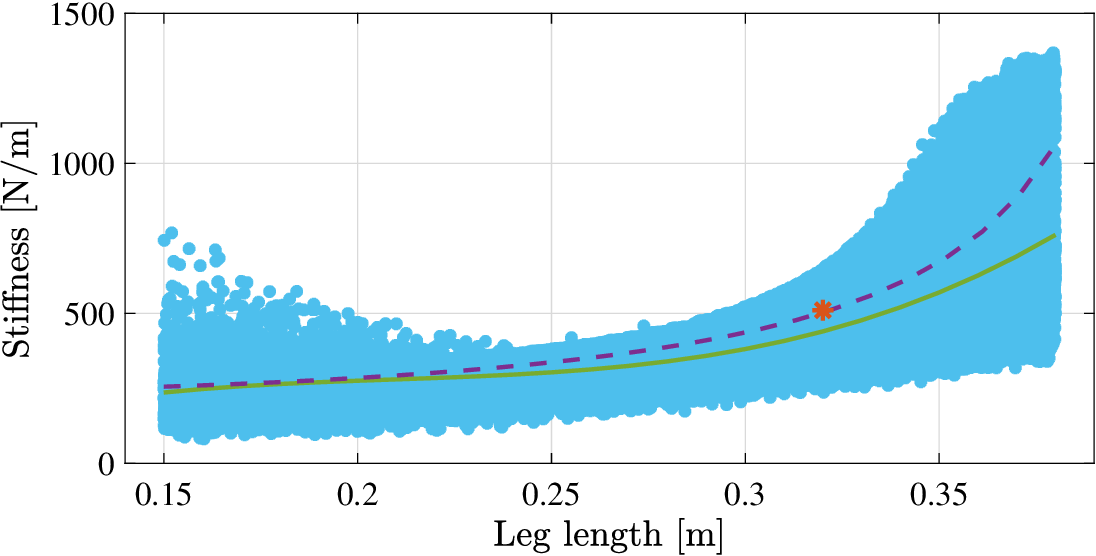}
		\caption{The equivalent spring stiffness regarding leg length. The raw data (marked by the blue colour) is sampled from a pre-designed workspace. The green solid curve shows the fitness result, while the purple dash curve shows varying stiffness regarding the vertical height. The red star in the purple curve marks the equivalent stiffness when the robot stands in the homing pose.}
		\label{fig:stiffness_mapping}
	\end{figure}
	
	\subsubsection{Nonlinear stiffness mapping}\label{sec:nonlinear_stiffness}
	Since the Jacobian ($\mathbf{J}^j$ in \eqref{potential_slip_match}) is configuration dependent, the equivalent spring constant ($k_s^i$) of the TD-aSLIP will change as the robot's configuration varies. Further analysis reveals that the varying $k_s^i$ is determined by a nonlinear function regarding multiple variables, including the body orientation and the legged joint angles. On the other hand, after mapping the quadrupedal body posture to the SLIP's body posture, the SLIP's leg length (i.e., $||\bm{l}_1||_2$ and $||\bm{l}_2||_2$ in \eqref{equ:assp1_slip2}) is also determined by joint angles of quadrupedal legs. Thus, we here associate the equivalent spring constant $k_s^i$ with the leg length, instead of the quadrupedal joint angles.  
	
	To draw the connection between the spring constant and the leg length, we sample data from the refined workspace. Assuming the four legs are located in the homing position (determined by the body size), we sample the body position from $[-0.1,0.1]\,$m, $[-0.1, 0.1]\,$m, $[0.15,0.37]\,$m, and the body inclination from $[-30^{\circ},30^{\circ}]$, $[-30^{\circ}, 30^{\circ}]$, $[-30^{\circ},30^{\circ}]$. Fig.~\ref{fig:stiffness_mapping} plots the resultant spring constant w.r.t.\ the leg length\footnote{The spring stiffness of the hip, thigh, and calf springs are $0\,$N/rad, $6\,$N/rad, and $12\,$N/rad, respectively, which are used in hardware experiments. }.
	
	As can be seen in Fig.~\ref{fig:stiffness_mapping}, the spring constant varies even with the same leg length, since one leg length can be computed from multiple body postures, resulting in multiple joint angles and Jacobians. To extract the relationship between spring constant and leg length, we use the linear regression method to achieve polynomial fitness. For simplification, a cubic polynomial is used. The fitted result is plotted by the green curve in Fig.~\ref{fig:stiffness_mapping}.
	
	In addition, the stiffness variation regarding the height variation is also reported, plotted by the purple dash curve\footnote{The robot stands in place with zero inclination angles in this scenario. Therefore, the leg length coincides with the body height.}. Compared with this situation, the sampling-based fitness also considers the configuration variation, which is thus closer to the real case. In the following, we will use the polynomial fitness when generating compliant motions. The generated motion with varying stiffness is discussed in Section~\hyperlink{motion_with_varying_stiffness}{VII-B}.
	
	\section{Dual-layered singularity-free motion generation} \label{sec_motion_planner}
	Given the task requirement, we first generate a coarse trajectory with the TD-aSLIP model. Then, warm-started by the raw trajectory, we optimize both the Cartesian and joint movements. In particular, the quaternion representation of the body rotation is incorporated in the second-layer TO to avoid singularities.

	\subsection{SLIP-driven motion optimization}
	
	\begin{figure}
		\centering
		\includegraphics[width=\columnwidth]{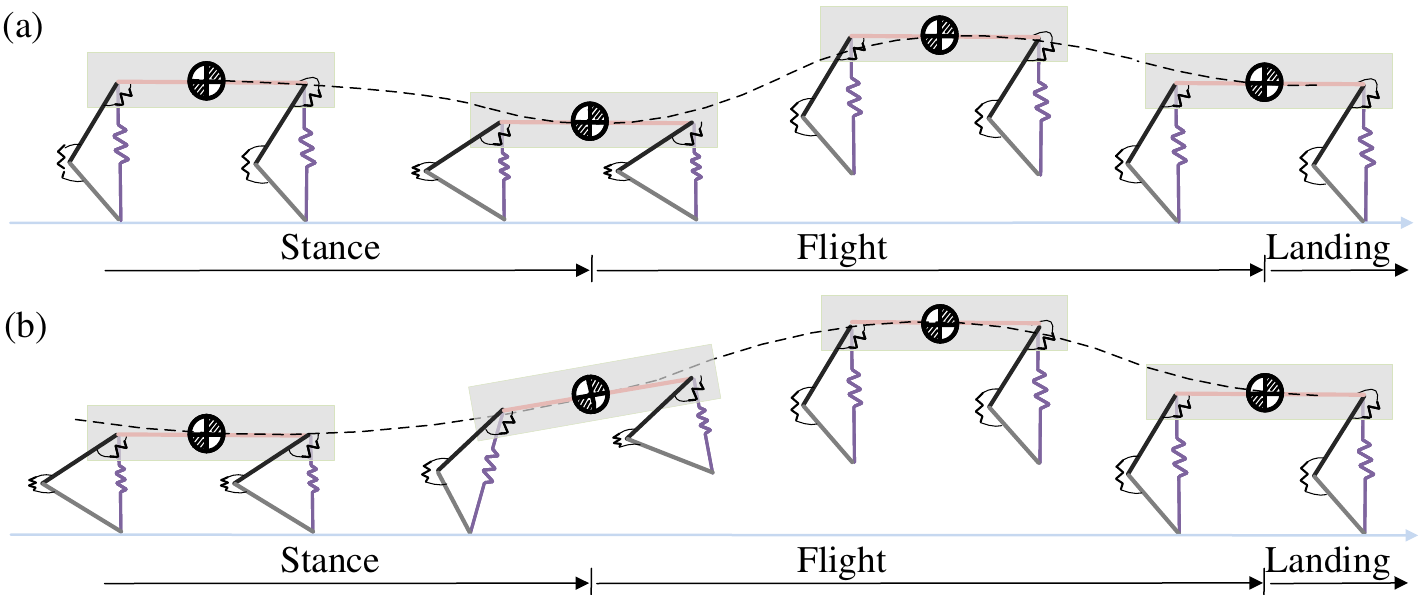}
		\caption{Contact sequence for (a) pronking  and (b) froggy jumping .}
		\label{fig:jumping_modal}
	\end{figure}
	Versatile explosive motion can be generated by defining different contact sequences. Figure~\ref{fig:jumping_modal} (a) and (b) 
	depict the pronking and froggy jumping motion, respectively.
	Taking the quadrupedal pronking, for example, we state the TO formulation for generating the coarse reference, given a desired jumping distance. As illustrated in Fig.~\ref{fig:jumping_modal}(a), we here focus on the body motion, which is divided into a stance and a flight phase, with all feet lifting off and touching down simultaneously. Assuming $N_t$ knots with $N_s$ knots for the stance phase (each knot lasts $t_s$) and $N_f$ knots for the flight phase (each knot lasts $t_f$), we then solve  the following problem,
	\begin{subequations}
		\begin{align}
			{\mathop{\arg\!\!\!\!\min\limits_{\M{F}_a^1, \M{F}_a^2, \ddot{\mathbf{X}}, \V t}} } \quad &J_\text{cost} \label{eq:kino_SLIP_cost}, 
			\\
			\text{s.t.} \quad \!\!\! 
			&\text{Waypoints constraints},
			\label{eq:waypoint}
			\\     
			&\text{State transition}, 
			\label{eq:kino_SLIP_xk+1}\\
			&\text{Dynamics constraints},
			\label{eq:kino_SLIP_dyn} \\			
			&\text{State boundary},
			\label{eq:state_boundary_constraint} \\
			&\text{Kinematics reachability},
			\label{eq:reeachbility_constrain} \\		
			&\text{GRF constraints},
			\label{eq:kino_SLIP_dyn_con} \\   
			&\text{Friction cone constraints} 
			\label{eq:kino_SLIP_friction_con}, \\
			&\text{Time step constraints} , 
			\label{eq:kino_SLIP_dt}    		 
		\end{align}
	\end{subequations}
	where the decision variables contain $\ddot{\mathbf{X}}$, $\mathbf{F}_a^1$, $\mathbf{F}_a^2$, and $\V {t}$. 
	$\ddot{\mathbf{X}} \in {\mathbb{R}^{6\times N_t}}$ comprises the 3D linear acceleration and the angular acceleration, 
	$\mathbf{F}_a^1 $ and $\mathbf{F}_a^2$ ($\in \mathbb{R}^{3\times N_s}$) denote actuation forces in stance, 
	$\bm{t} = [t_s,t_f]^{\text{T}}$ are the step sizes, $N_t \!=\! N_s\!+\!N_f$. 
	\subsubsection{Cost function}
	The cost function in (\ref{eq:kino_SLIP_cost}) is defined as
	\begin{subequations}
		\begin{align}		
			\label{eq:cost_slip_to}	
			&\quad \!\! J_\text{cost}\!=\!J_\text{stance}\!+\!J_\text{peak}\!+\!J_{\text{flight}}\!+\!{J_\text{land}} \!+\!J_{t},
			\\
			&\text{with} \nonumber \\
			& J_\text{stance} =\!\!  \sum_{k=1}^{N_s} (w_F^1\Vert \mathbf{F}^1_{a(k)}\Vert^2\!+\!w_F^2\Vert \mathbf{F}^2_{a(k)}\Vert^2\!+\!{w_{a}}\Vert \ddot{\mathbf{X}}_{(k)}\Vert^2 \nonumber\\&\qquad \qquad \quad \!+\!{w_{p}}\Vert{\mathbf{X}}_{(k)}\Vert^2\!+\!{w_{v}}\Vert\dot{\mathbf{X}}_{(k)}\Vert^2),\\   
			&\quad \!\!\! {J_\text{peak} = w_m\Vert \M{X}_{(N_s\!+\!N_f/2)} - \M{X}_{p}^r\Vert^2,}\\
			&\quad \!\!\!\!  J_{\text{flight}} = \sum_{k=N_s+1}^{N_t} \mathit{w}_{z}\Vert \bm{X}_{z(k)} - \bm{z}_{\text{takeoff}}\Vert^2,\\             
			&\quad \!\! {J_\text{land} =w_l\Vert \M{X}_{(N_t)} - \M{X}_{l}^r\Vert^2,}\\   
			&\qquad \!\!\!  J_{t} =  \mathit{w}_{t}\Vert \bm{t} - \bm{t}^{r}\Vert^2,
		\end{align}
	\end{subequations}
	where $J_\text{stance}$ penalizes control inputs (i.e., $\mathbf{F}^1_{a(k)}$ and $\mathbf{F}^2_{a(k)}$), body position and body rotation angle (${\mathbf{X}} \in {\mathbb{R}^{6\times N_t}}$), linear body velocity and angular velocity ($\dot{\mathbf{X}} \in {\mathbb{R}^{6\times N_t}}$) and CoM acceleration in stance phase. 
	$J_\text{peak}$ penalizes deviations from the reference peak state (${\mathbf{X}_p^r} \in {\mathbb{R}^{6}}$) in the middle of the flight,  $J_\text{flight}$ penalizes deviations from the reference shooting height ($z_{\text{takeoff}}$) during the flight phase with $\bm{X}_{z}\in {\mathbb{R}^{N_t}}$ being the CoM height,
	$J_\text{land}$ minimizes deviations from the reference landing state (${\mathbf{X}_l^r} \in {\mathbb{R}^{6}}$), $J_{t}$ penalizes deviations from the reference time steps ($\bm{t}^{r} = [t_s^{r}, t_f^{r}]^{\text{T}}$), , $w_F^1$, $w_F^2$, $w_{a}$, $w_p$, $w_v$, $w_m$, $w_{l}$, $w_{t}$ are positive weights.

	\subsubsection{Feasibility constraints} To ensure the feasibility, following constraints are considered in this TO formulation.
	
	The constraints in \eqref{eq:waypoint} are imposed to control the pronking motion to pass through three waypoints, including the initial status, take-off status, and landing status. That is,
	\begin{subequations}
		\label{jumping_initial_conditionx}
		\setlength{\abovedisplayskip}{3pt}
		\setlength{\belowdisplayskip}{3pt}
		\begin{align}
			&\quad \!\! \mathbf{X}_{(1)} = \V{X}_0,   \quad \!\! \dot{\mathbf{X}}_{(1)} = \dot{\V{X}}_0, \label{eq:kino_SLIP_velini}\\ 
			&\quad \!\! {(1\!-\!\xi) \M{X}_t^r \!\leq \!\M{X}_{(N_s)} \!\leq \!(1\!+\!\xi)\M{X}_t^r,}
			\label{eq:kino_SLIP_takeoff}\\             
			&\quad \!\! {(1\!-\!\xi)\M{X}_l^r \!\! \leq \!\!\M{X}_{(N_t)} \!\! \leq \!\! (1\!+\!\xi)\M{X}_l^r,} \label{eq:kino_SLIP_finalx} 
		\end{align}
	\end{subequations}	
	where  $\V{X}_0 \in \mathbb{R}^{6}$ and $\dot{\V{X}}_0 \in \mathbb{R}^{6}$ are the initial pose and velocity, ${\mathbf{X}_t^r} \in \mathbb{R}^{6}$ and ${\mathbf{X}_l^r} \in \mathbb{R}^{6}$ are the reference trunk motion at the take-off and landing moments, respectively. $\xi \in \mathbb{R}$ is a slack variable that is turned by hand.	
	
	When starting from the homing pose with zero velocity (as illustrated in Fig.~\ref{fig:quadrupedal_slip_model}(b)), we have
	\begin{equation}
		\label{jumping_initial_condition}
		\setlength{\abovedisplayskip}{3pt}
		\setlength{\belowdisplayskip}{3pt}
		\begin{aligned}
			{\V{X}_0\!=\!\!
				\left[\!\! \begin{array}{ccc}
					{0, 0, z_0, 0,0,0}
				\end{array} 
				\!\! \right]^{\text{T}}\!\!,} \quad 
			{\dot{\V{X}}_0\!=\!\!
				\left[\!\! \begin{array}{ccc}
					0,0, 0,0,0,0
				\end{array} 
				\!\! \right]^{\text{T}}\!\!},
		\end{aligned}
	\end{equation}
	with $z_0$ being the homing height.
	
	The state transition between neighboring nodes is achieved by \eqref{eq:kino_SLIP_xk+1}. For CoM position, the state transition is realized by the Taylor integration, following
	\begin{equation}
		\setlength{\abovedisplayskip}{3pt}
		\setlength{\belowdisplayskip}{3pt}
		\begin{aligned}
			\mathbf{X}^{1:3}_{(k+1)} = \mathbf{X}_{(k)}^{1:3} + \dot{\mathbf{X}}^{1:3}_{(k)}dt + \frac{1}{2}\ddot{\mathbf{X}}^{1:3}_{(k)}dt^2,
		\end{aligned}
	\end{equation}
	with $dt \in \{dt_s, dt_f\}$ according to the contact phase.
	
	However, for body rotation, we have 
	\begin{equation}
		\setlength{\abovedisplayskip}{3pt}
		\setlength{\belowdisplayskip}{3pt}
		\begin{aligned}
			\mathbf{X}^{4:6}_{(k+1)} = {\mathbf{X}}^{4:6}_{(k)} + \mathbf{T}_r\dot{\mathbf{X}}^{4:6}_{(k)}dt,
		\end{aligned}
	\end{equation}
	with where $\mathbf{X}^{4:6}_{(k)}=[\phi,\theta,\psi]^{\text{T}}$ separately comprise roll, pitch, and yaw angles\footnote{Here, we use the Euler angle to characterize the body rotation since it is easy to impose constraints on the rotational status, as done in Fig.~\eqref{eq:kino_SLIP_takeoff}$\sim$\eqref{eq:kino_SLIP_finalx}.}, $\dot{\mathbf{X}}^{4:6}_{(k)}=[\omega_x, \omega_y, \omega_z]^{\text{T}}$ comprise the angular velocities in the body frame.
	$\mathbf{T}_{r}\in \mathbb{R}^{3\times3}$ transforms the angular velocities in the body frame to the the change rate of the Euler angles, which is determined as 
	\begin{equation}\label{body_anguler_velo_tran}
		\mathbf{T}_{\text{r}} 
		=
		\begin{bmatrix}
			1 & \sin\phi \tan\theta & \cos\phi \tan\theta \\
			0 & \cos\phi & -\sin\phi \\
			0 & {\sin\phi}/{\cos\theta} & {\cos\phi}/{\cos\theta}
		\end{bmatrix}.
	\end{equation}

	For linear velocity and angular velocity, we have
	\begin{equation}
		\setlength{\abovedisplayskip}{3pt}
		\setlength{\belowdisplayskip}{3pt}
		\begin{aligned}
			\dot{\mathbf{X}}_{(k+1)} =  \dot{\mathbf{X}}_{(k)} + \ddot{\mathbf{X}}_{(k)}dt.
		\end{aligned}
	\end{equation}

	Eq.~\eqref{eq:kino_SLIP_dyn} ensures that the generated motion obeys the TD-aSLIP dynamics, i.e., \eqref{equ:trunk_slip_dual_legs} in stance and \eqref{equ:trunk_slip_dual_legs_flight} in flight.
	
	Eq.~\eqref{eq:state_boundary_constraint} ensures the robot states are within a feasible range, following inequality constraints defined by
	\begin{align}			
		\underline{\mathbf{X}} \leq \mathbf{X}_{(k)} \leq \overline{\mathbf{X}}, \quad \underline{\dot{\mathbf{X}}} \leq \dot{\mathbf{X}}_{(k)} \leq \overline{\dot{\mathbf{X}}},
	\end{align}
	Furthermore, in the stance phase, the leg length is constrained by \eqref{eq:reeachbility_constrain}, following
	\begin{align}			
		\underline{l} \leq ||\operatorname{Fk}(\mathbf{X}_{(k)},\mathcal{B}_i) - \bm{p}_i)||_2 \leq \overline{l}.
	\end{align}
	with $\operatorname{Fk}(\mathbf{X}_{(k)},\mathcal{B}_i)$ denotes the forward kinematics operation that calculates the hip position, $\mathcal{B}_i$ denote the $i$-th hip position in the body framework, $\bm{p}_i$ ($i \in \{1,2\}$) represent the homing leg positions of the TD-aSLIP, as illustrated in Fig.~\ref{fig:quadrupedal_slip_model}(b).
	
	Eq.~\eqref{eq:kino_SLIP_dyn_con} restricts the GRF and \eqref{eq:kino_SLIP_friction_con} considers the friction cone constraint during the stance phase, defined by
	\begin{subequations}
		\begin{align}
			0 \! &\leq \! \bm{F}^{i}_{z(k)} \!\!+\!\! {\bm{F}^{i}_{s,z}(\mathbf{X}_{(k)})} \! \leq \! \overline{F}_z, \quad \!\!\! i \! \in \!\! \{1,2\},
			\label{eq:forcez_con} \\   
			-\mu &\leq {\ddot{\bm{X}}_{\lambda(k)}}/({\ddot{\bm{X}}_{z(k)}\!+\!g}) \leq \mu,  \quad \!\!\! \lambda \!\! \in \!\! \{x,y\},
			\label{eq:frictionz_con}
		\end{align}
	\end{subequations}
	where $\bm{F}^{i}_{z} \in \mathbb{R}^{N_s} $ and ${\bm{F}^{i}_{s,z}} \in \mathbb{R}^{N_s}$ separately represent the vertical actuation force and vertical spring force of the $i$-th leg, $\ddot{\bm{X}}_{x}\in {\mathbb{R}^{N_t}}$ and $\ddot{\bm{X}}_{y}\in {\mathbb{R}^{N_t}}$ separately represent the forward and lateral CoM acceleration, $g$ is the vertical gravitational acceleration, and $\mu$ is the friction coefficient.
	
	Eq.~\eqref{eq:kino_SLIP_dt} restricts the variation of the step size, following 
	\begin{align}			
		\underline{\bm{t}} \leq \bm{t} \leq \overline{\bm{t}}.
	\end{align}
	

	\textit{Remark 1}: Through modulating contact sequence and waypoints, other jumping motions such as froggy jumping (of which the key contact sequences are illustrated in Fig.~\ref{fig:jumping_modal}(b)) can also be generated by this SLIP TO. Also, in our formulation, omnidirectional jumps, including hop-turn, can also be realized, which will be demonstrated in the following sections.

	\subsection{Kinodynamics optimization with singularity avoidance}
	The first layer generates a coarse trajectory for a desired task without ensuring a feasible joint movement. Also, a coarse Euler angle profile is obtained to represent body rotation, which may face the Gimbal lock while performing highly dynamic motions\footnote{In practice, we found the first-layer SLIP TO could find a feasible solution when experiencing large rotation, such as with a $\pm90^\circ$ roll, pitch, or yaw.}. 
	In the second layer, we address these issues with kinodynamics optimization, which is defined as
	\begin{subequations}
		\begin{align}
			{\mathop{\arg \min\limits_{\M{F}_a^1, \M{F}_a^2, \ddot{\mathbf{X}}, \V t, {\M{q}}}} } \quad &\!\!\!J_\text{cost} \label{eqx:kino_SLIP_cost}, 
			\\
			\text{s.t.} \quad  
			&\!\!\!\text{Waypoint constraints}, \label{eqx:kino_SLIP_velini}\\ 
			&\!\!\!\text{State transition}, 
			\label{eqx:kino_SLIP_finalz}
			\\    
			&\!\!\!\text{Dynamics constriants}
			\label{eqx:kino_SLIP_dt}\\     
			&\!\!\!\text{State boundary},
			\label{eqx:state_con}\\
			&\!\!\!\text{Kinematic reachability}, 
			\label{eqx:kine_reach_con}\\   
			&\!\!\!\text{GRF constraints}, 
			\label{eqx:kino_grf_con}\\       
			&\!\!\! \textbf{Contact constraints}, \label{eqx:kino_contact_con} \\
			&\!\!\! \textbf{Torque constraints}, \label{eqx:kino_torque_con} \\
			&\!\!\! \text{Time step constraints}, \label{eqx:kino_time_step_con}
		\end{align}
	\end{subequations}
	where the joint angles $\M{q} \!\in \!\mathbb{R}^{6\times N_t}$ are added into the decision variables,`\textbf{Contact constraints}' in \eqref{eqx:kino_contact_con} and `\textbf{Torque constraints}' in \eqref{eqx:kino_torque_con} are introduced to ensure the dynamic feasibility.
	
	\subsubsection{Cost function}
	The cost function in (\ref{eqx:kino_SLIP_cost}) is defined following \eqref{eq:cost_slip_to}. In addition, we consider tracking the reference quaternion, reference angular velocity, and the homing joint angles, by adding the following item ($J_\text{add}$):
	\begin{align} \label{eq:cost_kino_add}			   
		&J_\text{add} = \!\! \!\sum_{k=1}^{N_t} \!\!(w_Q (d_{\text{euclid}} (\mathbf{X}^{4:7}_{(k)}, \mathbf{Q}_{(k)}^r))^2 \!\!+\!\! w_{\omega}\Vert \dot{\mathbf{X}}^{4:6}_{(k)} \!-\! (\dot{\mathbf{X}}^{4:6}_{(k)})^r \Vert^2) \nonumber \\
		&\qquad \!+\! \sum_{k=1}^{N_s} (w_q\Vert \M{q}_{(k)} - \bm{q}_0\Vert^2),
	\end{align}
	where $\mathbf{X}^{4:7} \in \mathbb{R}^{4\times N_t}$ and $\mathbf{Q}^r \in \mathbb{R}^{4\times N_t}$ are the optimized and reference quaternion. $\dot{\mathbf{X}}^{4:6}\in \mathbb{R}^{3\times N_t}$ and  $(\dot{\mathbf{X}}^{4:6})^r \in \mathbb{R}^{3\times N_t}$ are the optimized and reference angular velocities. 
	$\bm{q}_0 \in \mathbb{R}^6 $ is the initial joint angle corresponding to homing pose. $w_Q$, $w_{\omega}$ and $w_q$ are positive weights.
	
	In this work, we use the Euclidean distance to measure the difference between two quaternions (e.g., $\bm{Q}_1$ and $\bm{Q}_2$)\footnote{A more accurate metric is to use the Geodesic distance, which, however, results in a non-convex and non-smooth cost function.}. Considering that quaternions have the “double-cover” property, we compute $d_{\text{euclid}}$ in \eqref{eq:cost_kino_add} as
	\begin{align}
		d_{\text{euclid}}(\bm{Q}_1, \bm{Q}_2) = \min\left( \left\| \bm{Q}_1 - \bm{Q}_2 \right\|_2,\; \left\| \bm{Q}_1 + \bm{Q}_2 \right\|_2 \right).
	\end{align}
	
	In \eqref{eq:cost_kino_add}, the reference angular velocity ($(\dot{\mathbf{X}}^{4:6})^r$) and quaternion ($\mathbf{Q}^r$) are generated from first-layer optimization. Specifically, the reference quaternion at the $k$-th step ($\mathbf{Q}^r_{(k)}$) is computed as
	\begin{equation}\label{qh_ref}
		\mathbf{Q}^r_{(k)} \!=\! \mathbf{Q}^r_{(k-1)}\!+\!\dot{\mathbf{Q}}^r_{(k-1)}dt  \!= \!\mathbf{Q}^r_{(k-1)}\!+\!\frac{1}{2}\bm{\Omega}(\boldsymbol{\omega}) \mathbf{Q}^r_{(k-1)}dt,
	\end{equation}
	where $dt$ is the time interval, the matrix $\bm{\Omega}(\boldsymbol{\omega})$ is determined by the reference angular velocity ($[\omega_x^r,\omega_y^r,\omega_z^r]^{\text{T}}$) obtained by the first-layer TO, which follows
	\begin{equation}\label{quaterion_velo_update}
		\bm{\Omega}(\boldsymbol{\omega}) =
		\begin{bmatrix}
			0 & -\omega_x^r & -\omega_y^r & -\omega_z^r \\
			\omega_x^r & 0 & \omega_z^r & -\omega_y^r \\
			\omega_y^r & -\omega_z^r & 0 & \omega_x^r \\
			\omega_z^r & \omega_y^r & -\omega_x & 0
		\end{bmatrix}.
	\end{equation}

	
	\subsubsection{Kinodynamics constraints}
	Constraints are defined following the first-layer formulation. Here, we focus on the constraints regarding joint motion and quaternion variation.
	
	`{Waypoint constraints}' are enhanced with additional constraints. Taking the initial state as an example, we have
	\begin{subequations}
		\begin{align}
			\mathbf{q}_{(1)} = \bm{q}_0.
			\label{eqx:kino_joint_angle_con}
		\end{align}
	\end{subequations}

	`{State transition}' is achieved by the Taylor integration, following the quaternion dynamics expressed in \eqref{qh_ref} and \eqref{quaterion_velo_update}. 

	The following `State boundary' conditions are added 
	\begin{align}
		\underline{\bm{q}}\leq \bm{q}_{(k)} \leq \overline{\bm{q}}, \quad \underline{\dot{\bm{q}}}\leq \dot{\bm{q}}_{(k)} \leq \overline{\dot{\bm{q}}}. 
		\label{eqx:kino_angle_con}
	\end{align}
	
	`\textbf{Contact constraints}' are added to prevent the leg movement in the stance phase, follows
	\begin{align}
		\operatorname{Fk}(\mathbf{X}_{(k)},\mathcal{B}_i, \M{q}_{(k)}) = \bm{p}_i, \quad i \in \{1,2\}. 
		\label{eqx:kino_SLIP_foot_con} 
	\end{align}
	
	The following `\textbf{Torque constraints}' are introduced 
	\begin{align}
		\underline{\bm{\tau}} \leq -\mathbf{J}^i(\mathbf{q}_{(k)}) (\mathbf{F}_{a(k)}^i + \mathbf{F}_{s}^i(\mathbf{X}_{(k)})) \leq \overline{\bm{\tau}}, 
		\label{eqx:kino_SLIP_torque_con} 
	\end{align}
	
	As a result, the optimal motion is obtained for a given task.    
	
	\textit{Remark 2}: The above dual-layer TO formulation releases the necessity of a hand-turned reference trajectory for achieving versatile motions. Instead, we only need to choose several waypoints. Furthermore, the generated quaternion profile would be directly passed into the torque controller, enabling singularity-free tracking.
	
	\section{Whole-body Compliance Control}\label{sec_compliance_control} 
	This section presents the motion tracking strategy, including feedfoward compensation (generating $\bm{\tau}^{ff}$) and feedback control (generating $\bm{\tau}^{fb}$). For feedfoward compensation, the quaternion MPC is formulated to generate the reference GRFs and joint torques. Then, an enhanced WBC is proposed to optimize the GRFs and joint torques, while explicitly considering the spring effect and safety constraints.

	\subsection{SRB-based quaternion MPC}
	The quaternion MPC is inspired by the work in (\cite{garcia2021time}), which is not our main contribution. So, we briefly introduce the basic idea.
	\subsubsection{SRB model with quaternion dynamics}\label{srb_quaternion}
	To simplify the formulation, we utilize the SRB model that neglects the leg mass and assumes a lumped mass in the body center. The body acceleration is then computed as
	\begin{subequations}
		\begin{align}
			\label{Eq:pcom}
			\ddot{\bm{c}} 	& = \sum_{i=1}^{n} \frac{\bm{f}_i}{m} + \bm{g},  \\
			\label{Eq:L}
			\dot{\bm{L}} 	& = \sum_{i=1}^{n} (\bm{p}^f_{i} - \bm{c}) \times \bm{f}_i, 
		\end{align}
	\end{subequations}
	where $\bm{f}_i \in \mathbb{R}^3$ is the GRF of the $i$-th leg, $\bm{p}^f_{i} \in \mathbb{R}^3$ is the $i$-th leg position,  $\bm{L} \in \mathbb{R}^3$ is the angular momentum around the CoM, and $n$ represents the number of legs.
	
	Eq.~\eqref{Eq:L} reveals that nonlinear angular momentum dynamics regarding $\bm{c}$ and $\bm{f}_i$. To address this nonlinearity, we introduce the global angular momentum ($\bm{L}_2 \in \mathbb{R}^3$) with respect to the world origin, determined as 
	\begin{equation}
		\label{Eq:L2_L}
		\bm{L}_2 = \bm{L} + m \bm{c} \times \dot{\bm{c}}.
	\end{equation}
	Then, the change rate of $\bm{L}_2$ is given by
	\begin{equation}
		\label{Eq:L2}
		\dot{\bm{L}}_2 = \sum_{i=1}^{n} \bm{p}^f_{i} \times \bm{f}_i + m \bm{c} \times \bm{g}.
	\end{equation}

	For rotational movements, we employ quaternions to circumvent the singularities of Euler angles and the inherent nonlinear complexities of rotation matrices. The quaternion dynamics is expressed as
	\begin{equation}\label{qhx}
		\dot{\bm{Q}} = \frac{1}{2} \bm{Q} \circ \boldsymbol{\omega} = \frac{1}{2} \mathbf{T}(\bm{Q}) \boldsymbol{\omega} \\,
	\end{equation}
	where $\mathbf{T}(\bm{Q})$ is 
	\begin{equation}
		\mathbf{T}(\bm{Q}) =
		\left[
		\begin{array}{ccc}
			-q_{x} & -q_{y} & -q_{z} \\
			q_{w} & -q_{z} & q_{y} \\
			q_{z} & q_{w} & -q_{x} \\
			-q_{z} & q_{x} & q_{w}
		\end{array}
		\right].
	\end{equation}
	
	By neglecting the Coriolis force, the angular velocity $\bm{\omega}$ is determined by
	\begin{equation}\label{ome}
		\boldsymbol{\omega} = \bm{\mathcal{{I}}}^{-1}\bm{L} =  (\textbf{R}\bm{\mathcal{{I}}}_b \textbf{R}^{\text{T}})^{-1}\bm{L}.
	\end{equation}
	
	Combing \eqref{qhx} and \eqref{ome}, the quaternion change is
	\begin{equation}\label{hdot}
		\dot{\bm{Q}} = \frac{1}{2} \mathbf{T}(\bm{Q})\textbf{R}\bm{\mathcal{{I}}}_{b}^{-1}\textbf{R}^{\text{T}}(\bm{L}_{2} - m \bm{\bm{c}} \times \dot{\bm{c}})  \\.
	\end{equation}

	\subsubsection{Linearized quaternion MPC}
	For MPC formulation, we define the state vector ($\bm{x}\in \mathbb{R}^{13}$) as
	\begin{equation}
		\bm{x} = 
		\begin{bmatrix}
			\bm{c}^{\text{T}},  
			\dot{\bm{c}}^{\text{T}}, 
			\bm{L}_{2}^{\text{T}},
			\bm{Q}^{\text{T}}
		\end{bmatrix}^{\text{T}}.
	\end{equation}
	%
	
	Assuming small variations between neighboring iterations, we apply a first-order Taylor expansion  of the nonlinear function \eqref{hdot} around the prior state $\bm{x}^{-}$, yielding the linearized expression:
	\begin{equation}\label{linargu}
		\dot{\bm{Q}} = \bm{\mathcal{F}}(\bm{x}^{-}) + \frac{\partial \bm{\mathcal{F}}}{\partial \bm{x}} \bigg|_{\bm{x} = \bm{x}^{-}} \left( \bm{x} - \bm{x}^{-} \right),
	\end{equation}
	where $\bm{\mathcal{F}}(\bm{x}) = \dot{\bm{Q}}$ represent the expression in \eqref{hdot}, $\frac{\partial \bm{\mathcal{F}}}{\partial \bm{x}}$ refers to the partial derivative.
	
	Then, the SRB dynamics is expressed as 
	\begin{equation}
		\dot{\bm{x}} = \mathbf{A}(\bm{x}^{-})\bm{x} + \mathbf{B}(\bm{x}^{-})\bm{u} + \mathbf{C}(\bm{x}^{-}),
	\end{equation}
	where $\mathbf{A}(\bm{x}^{-})$, $\mathbf{B}(\bm{x}^{-})$, and $\mathbf{C}(\bm{x}^{-})$ are derived following the dynamics defined in Section~\hyperlink{srb_quaternion}{VI-A1}. Please refer to (\cite{garcia2021time}) for further details.
	
	The continuous-time system above is then discretized using the matrix exponential method, with a time interval corresponding to the MPC step size $\Delta t$. The resulting discretized form is given by
	\begin{align}
		\mathbf{x}[k+1] = \mathbf{A}_d[k]\mathbf{x}[k] + \mathbf{B}_d[k]\mathbf{u}[k] + \mathbf{C}_d[k],
	\end{align}
	with
	\begin{align}	&\mathbf{A}_d[k] = \exp(\mathbf{A}(\mathbf{x}^{-}_{(k)})\Delta t), \\
		&\mathbf{B}_d[k] = \mathbf{A}^{-1}(\mathbf{x}^{-}_{(k)}) \left( \exp(\mathbf{A}(\mathbf{x}^{-}_{(k)})\Delta t) - {\mathbf{{I}}} \right) \mathbf{B}(\mathbf{x}^{-}_{(k)}), \\
		&\mathbf{C}_d[k] = \mathbf{A}^{-1}(\mathbf{x}^{-}_{(k)}) \left( \exp(\mathbf{A}(\mathbf{x}^{-}_{(k)})\Delta t) - {\mathbf{{I}}} \right) \mathbf{C}(\mathbf{x}^{-}_{(k)}),
	\end{align}
	where $\mathbf{x}^{-}_{(k)}$ is the prior state, and $\exp(\cdot)$ represents the matrix exponential of matrix $(\cdot)$.

	With the above linearization, we then adopt a standard MPC scheme (\cite{di2018dynamic}) to calculate the desired GRFs ($\bm{f}^{\text{mpc}}_r \in \mathbb{R}^{12}$). After solving the GRFs, the feedforward torques {($\bm{\tau}^\text{mpc} \in \mathbb{R}^{12}$) for GRF compensation are computed as
		\begin{subequations}
			\begin{align}
				\bm{\tau}^\text{mpc} =  - \mathbf{J}_c^{\text{T}} \bm{f}^{\text{mpc}}_r.\label{ff_torque_algorithm_sub}
			\end{align}
		\end{subequations}
		with $\mathbf{J}_c \in \mathbb{R}^{12\times12}$ comprising the contact Jacobian.

		\subsection{WBC with parallel compliance}
		The above MPC scheme generates the raw GRF and torque command for accomplishing a given task. Since it only considers the simplified dynamics, the above scheme may not work for highly dynamic motions. For better tracking, WBC could be employed. However, the traditional WBC seldom addresses the parallel compliance introduced by the mechanical design. Furthermore, the safety constraints are usually ignored.
		
		Taking the raw GRF and torque profiles generated by the above MPC as initial guesses, we here generate the optimal torque profiles with a whole-body compliant model that explicitly considers parallel compliance. Furthermore, we explicitly consider torque limits and other safety constraints, such as collision avoidance. As a result, we propose the following quadratic programming (QP) formulation,
		\begin{subequations}
			\begin{align}
				\min_{\bm{\delta}_{f}, \bm{\delta}_b} \quad &\bm{\delta}_{f}^\top \mathbf{W}_1 \bm{\delta}_{f} + \bm{\delta}_b^\top \mathbf{W}_2 \bm{\delta}_b+ \bm{\delta}_{\tau_j}^\top \mathbf{W}_3  \bm{\delta}_{\tau_j} \label{wbc_obj} \\
				\text{s.t.} \quad
				&\mathbf{S}_f \left( \mathbf{M} \ddot{\bm{q}} \!+\! \bm{B} \!+\! \bm{G}\!+\! {\bm{K}} \right) = \mathbf{S}_f \mathbf{J}_c^\top \bm{f}_r, \label{floating_dynamic_onstra} \\
				&\ddot{\bm{q}} =\ddot{\bm{q}}^{\text{cmd}} + 
				\begin{bmatrix}
					\bm{\delta}_b, \\
					\bm{0}_{12}
				\end{bmatrix}, \label{body_acceleration_tech}\\
				&\bm{f}_r = {\bm{f}^{\text{mpc}}_r} + \bm{\delta}_{f},\label{grf_tech}\\
				&\bm{\mathcal{W}} \bm{f}_r \geq 0,\label{grf_cons}\\
				&{\bm{\delta}_{\tau_j}} {=} {\bm{\tau}_j - \bm{\tau}^\text{mpc}}, \label{torque_devia} \\
				&{\bm{\tau}_j}
				{=} {\mathbf{S}_j(\mathbf{A}\ddot{\bm{q}} \!+\! \bm{B} \!+\!  \bm{G} \!+\! \bm{K}\!-\!\mathbf{J}_c^\top  \bm{f}_r),}\\
				&{\underline{\bm{\tau}}_j} {\leq}  {\bm{\tau}_j}  {\leq} {\overline{\bm{\tau}}_j}, \label{torque_limits}  \\
				&{\dot{\bm{h}}(\bm{\delta}_{b}) + \lambda {\bm{h}}(\bm{\delta}_{b})}  {\geq} {0.} \label{cbf_safety}
			\end{align}
		\end{subequations}
		
		\subsubsection{Cost function} \eqref{wbc_obj} defines the cost function for the whole-body control optimization. The aim of this WBC is to penalize the GRF deviation ($\bm{\delta}_{f} \in \mathbb{R}^{12}$), body acceleration deviation ($\bm{\delta}_b \in \mathbb{R}^{6}$), and joint torque deviation ($\bm{\delta}_{\tau_j} \in \mathbb{R}^{12}$), while obeying the full-body dynamics and other feasibility constraints. $\mathbf{W}_1$, $\mathbf{W}_2$, and $\mathbf{W}_3$ are the weight matrices.
		
		\subsubsection{Feasibility constraints}
		Eqs.~\eqref{floating_dynamic_onstra}~to~\eqref{grf_tech} capture the floating-based dynamics where $\mathbf{M}$, $\bm{B}$, $\bm{G}$, and $\bm{K}$ separately denote the inertial matrix, Coriolis, Gravity, and parallel spring torque, $\ddot{\bm{q}}\in \mathbb{R}^{18}$ and $\ddot{\bm{q}}^{\text{cmd}} \in \mathbb{R}^{18}$ are separately the generalized acceleration generated by WBC and the generalized acceleration generated by motion planner, $\bm{f}_r\in \mathbb{R}^{12}$ is the generated GRF, which is constrained by \eqref{grf_cons}. $\bm{\mathcal{W}}$ captures the GRF constraints. $\mathbf{S}_f$ selects the torque on the floating base.
		
		Eqs.~\eqref{torque_devia}~to~\eqref{torque_limits} impose constraints on the joint torque ($\bm{\tau}_j \in \mathbb{R}^{12}$) generated by WBC. $\mathbf{S}_j$ selects that joint torque.
		
		In \eqref{cbf_safety}, the function $\bm{h}(\bm{\delta}_{b})$ defines the safety set of the base motion\footnote{$\bm{h}(\bm{\delta}_{b}) \geq 0 $ means the current state is safe (\cite{ferraguti2022safety}).}. The expression in \eqref{cbf_safety} works as an {exponential control barrier function (CBF)} to restrict the system in the safety set (\cite{ames2019control}). Imagining that we require the body to stay above a certain height so as to avoid collisions with the ground, with \eqref{cbf_safety}, a safety constraint is then imposed to $\bm{\delta}_{b}$.
		
		As a result, the above WBC is formulated as a QP problem, computing the optimal feedforward torque.
		
		\textit{Remark 3}: {This WBC formulation draws inspiration from the work in (\cite{kim2019highly}). However, we do not use whole-body kinematics to adjust the joint reference. Instead, our WBC formulation explicitly limits the torque profile from the full-body compliant dynamics and considers the safety constraints. Differing from the CBF-MPC in (\cite{grandia2021multi}), our formulation can be run at a higher frequency, which is desired for highly dynamic motion. Furthermore, we can still make use of the prediction capability of MPC in this hierarchical control framework without significantly increasing the complexity.}

		\subsection{Hybrid torque controller}
		
		Finally, the computed joint angles ($\bm{q}_j^\text{cmd}$) and velocities ($\dot{\bm{q}}_j^\text{cmd}$) are sent to the joint-level controller. Combing with the whole-body torque commands, the final control inputs are:
		\begin{equation}
			\begin{aligned}
				\bm{\tau}^\text{cmd} &=  \bm{\tau}_{ff} +  \bm{\tau}_{fb} \\ &= \bm{\tau}_j + \mathbf{K}_p \left({\bm{q}}_j^\text{cmd} - {\bm{x}}_j\right) + \mathbf{K}_d \left({\dot{\bm{q}}_j^\text{cmd}} - \dot{\bm{q}}_j\right).
			\end{aligned}
		\end{equation} 
		where $\mathbf{K}_p$ and $\mathbf{K}_d$ are the PD gain matrices.
		


		\begin{figure}
			\centering
			\includegraphics[width=\columnwidth]{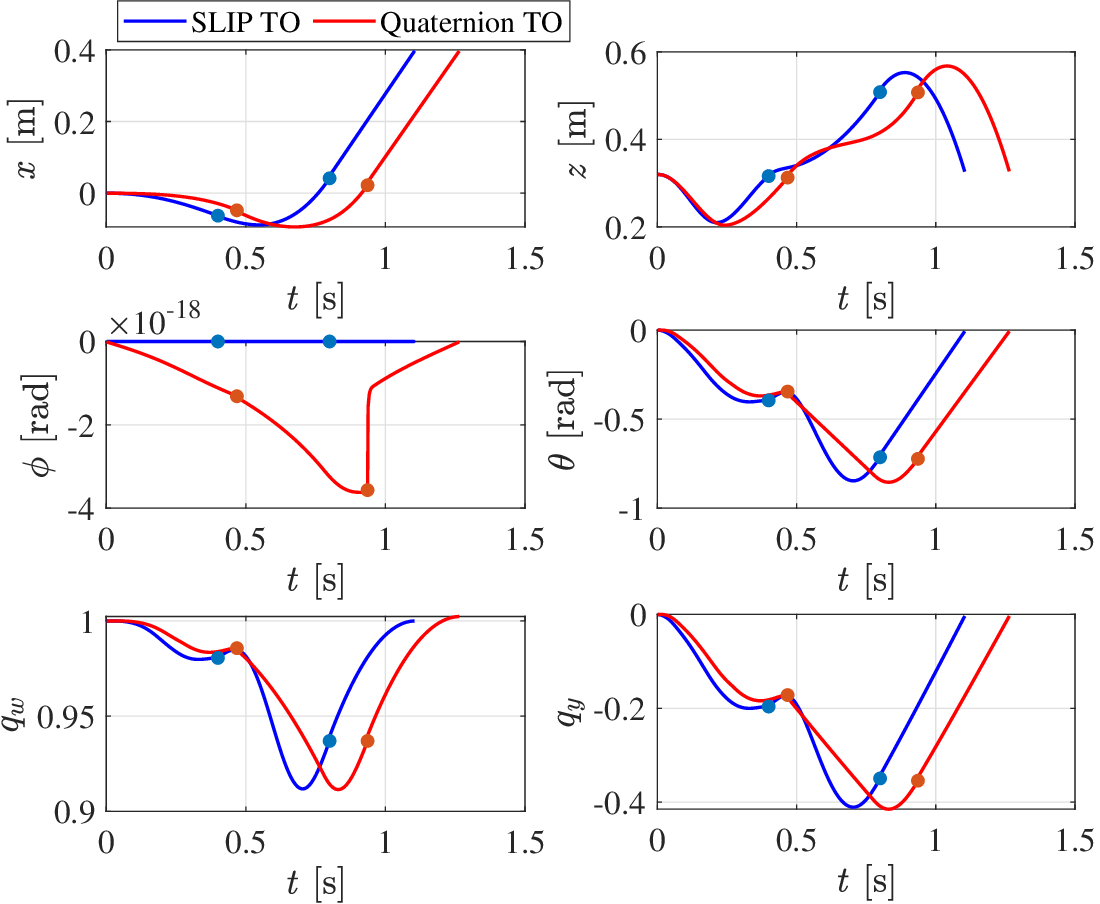}
			\caption{$40\,$cm forward froggy jumping trajectory generated by the proposed TO scheme. `SLIP TO' and `Quaternion TO' separately represent the first layer TO with SLIP dynamics and the dual-layer singularity-free TO. The two dots on each curve separately mark the transition from full-leg support to rear-leg support and the transition from rear-leg support to flight. From top to bottom, the CoM position, Euler angle and quaternion elements are presented, respectively. Since there is almost no lateral movement, and the roll and yaw angles are also close to zeros, we omit the lateral position ($y$), yaw angle ($\psi$), as well as $q_x$ and $q_z$, for brevity.}
			\label{fig:rigid_quadrupedal_motion_dual_layer_motion_generation}
		\end{figure}
		
		\begin{figure}
			\centering
			\includegraphics[width=\columnwidth]{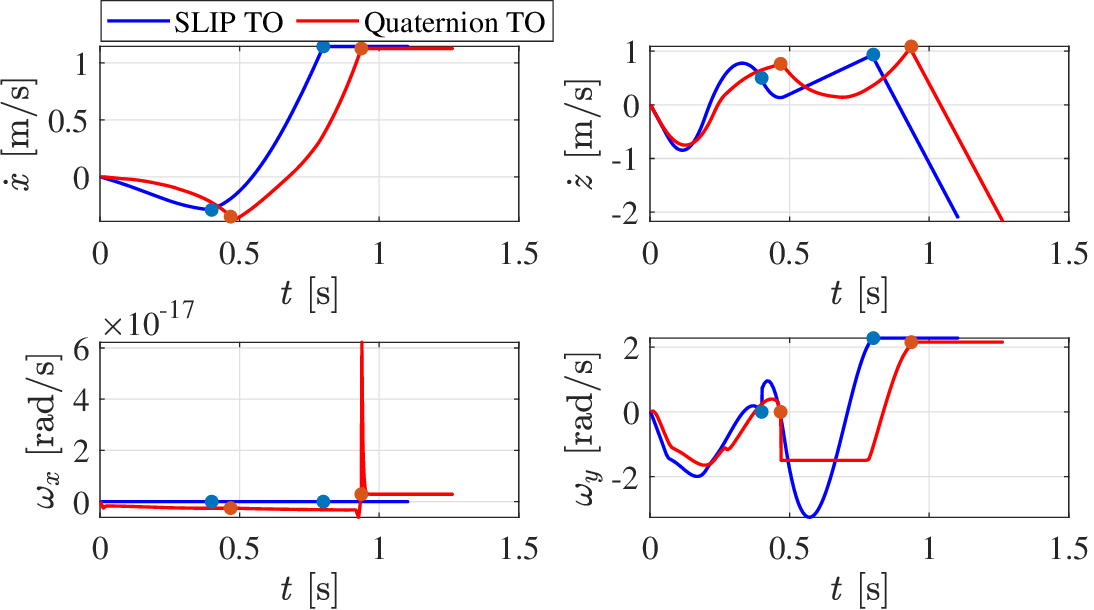}
			\caption{Velocity profiles for $40\,$cm forward froggy jumping. `SLIP TO' and `Quaternion TO' separately represent the first layer TO with SLIP dynamics and the dual-layer singularity-free TO. The top and bottom rows separately plot the translational velocity and rotational velocity. The two dots on each curve separately mark the transition from full-leg support to rear-leg support and the transition from rear-leg support to flight. The lateral CoM velocity ($\dot{y}$) and yaw velocity ($w_z$) are almost zeros, which are omitted here.}
			\label{fig:rigid_quadrupedal_motion_dual_layer_motion_generation_velo}
		\end{figure}
		
		\section{Simulation validation} \label{sec_evaluation}
		This section demonstrates the effectiveness of the proposed TD-aSLIP model and the dual-layer TO for motion generation. First, we validate the versatility of the novel TD-aSLIP model. Second, we investigate the dynamic motions with/without parallel compliance. Finally, we compare the novel TD-aSLIP model against a previous one, i.e., the dual-aSLIP without trunk rotation. For all tests, the (offline) dual-layer TO problem, i.e., the motion planner, is formulated via \texttt{CasADi} (\cite{andersson2019casadi}) with Python script, solved by \texttt{Ipopt} (\cite{wachter2006implementation}). The motion controller is run online, with `OSQP' (\cite{osqp}) being the QP solver.

		\subsection{Versatile motion generation} 
		\subsubsection{Dynamic motion generation via dual-layer TO}\label{sec:dual_layer_generation}
		By adding a rear-leg stance phase after the full-leg stance phase (see Fig.~\ref{fig:jumping_modal}(b)) and modulating the waypoints, the forward froggy jumping that requires a large body rotation can be generated. Take the rigid case for an example, Fig.~\ref{fig:rigid_quadrupedal_motion_dual_layer_motion_generation} and Fig.~\ref{fig:rigid_quadrupedal_motion_dual_layer_motion_generation_velo} display the optimal trajectories (including position and velocities) generated by the proposed dual-layer TO scheme (see plots with `Quaternion TO'). For comparison, we also display the coarse trajectories generated with the first-layer TO, denoted by `SLIP TO'.
		
		From Fig.~\ref{fig:rigid_quadrupedal_motion_dual_layer_motion_generation} and Fig.~\ref{fig:rigid_quadrupedal_motion_dual_layer_motion_generation_velo}, we can see that both the first-layer (denoted by `SLIP TO') and the dual-layer TO (denoted by `Quaternion TO') generate reference trajectories for the desired jumping task, i.e., realizing $40\,$cm forward jumping (see the top left figure in Fig.~\ref{fig:rigid_quadrupedal_motion_dual_layer_motion_generation}). Since more constraints, such as torque limits, are considered, the second-layer kinodynamics TO refined the coarse trajectory generated by the first-layer `SLIP TO'. In this case, the transitions between different contact phases happened later (marked by the solid dots in each curve) with the `Quaternion TO', meaning that the second-layer optimization extends the period of each contact phase to guarantee feasibility. 
		
		It is worth mentioning again that the second-layer TO generates the quaternion profile (see the red curves in the third row of Fig.~\ref{fig:rigid_quadrupedal_motion_dual_layer_motion_generation}) to represent the desired body rotation. In addition, by setting the waypoints properly, zero lateral movements, zero rolls, and zero yaws are achieved. For brevity, only the rotational angle ($\phi$ in Fig.~\ref{fig:rigid_quadrupedal_motion_dual_layer_motion_generation}) and the angular velocity ($\omega_x$ in Fig.~\ref{fig:rigid_quadrupedal_motion_dual_layer_motion_generation_velo}) around the $x-$ axis are shown, while the lateral CoM movement, yaw angle, and the elements of the quaternions $q_x$ and $q_z$ are omitted. 
		
		\subsubsection{Multi-modal explosive motion}
		In addition to froggy jumping, by changing the contact sequence and customizing the reference way-points, more versatile motions such as pronking and hop-turn could be generated. Fig.~\ref{fig:quadrupedal_motion_simu} visualizes the three explosive motions generated by the proposed dual-layer TO scheme.
		
		\begin{figure}
			\centering
			\includegraphics[width=\columnwidth]{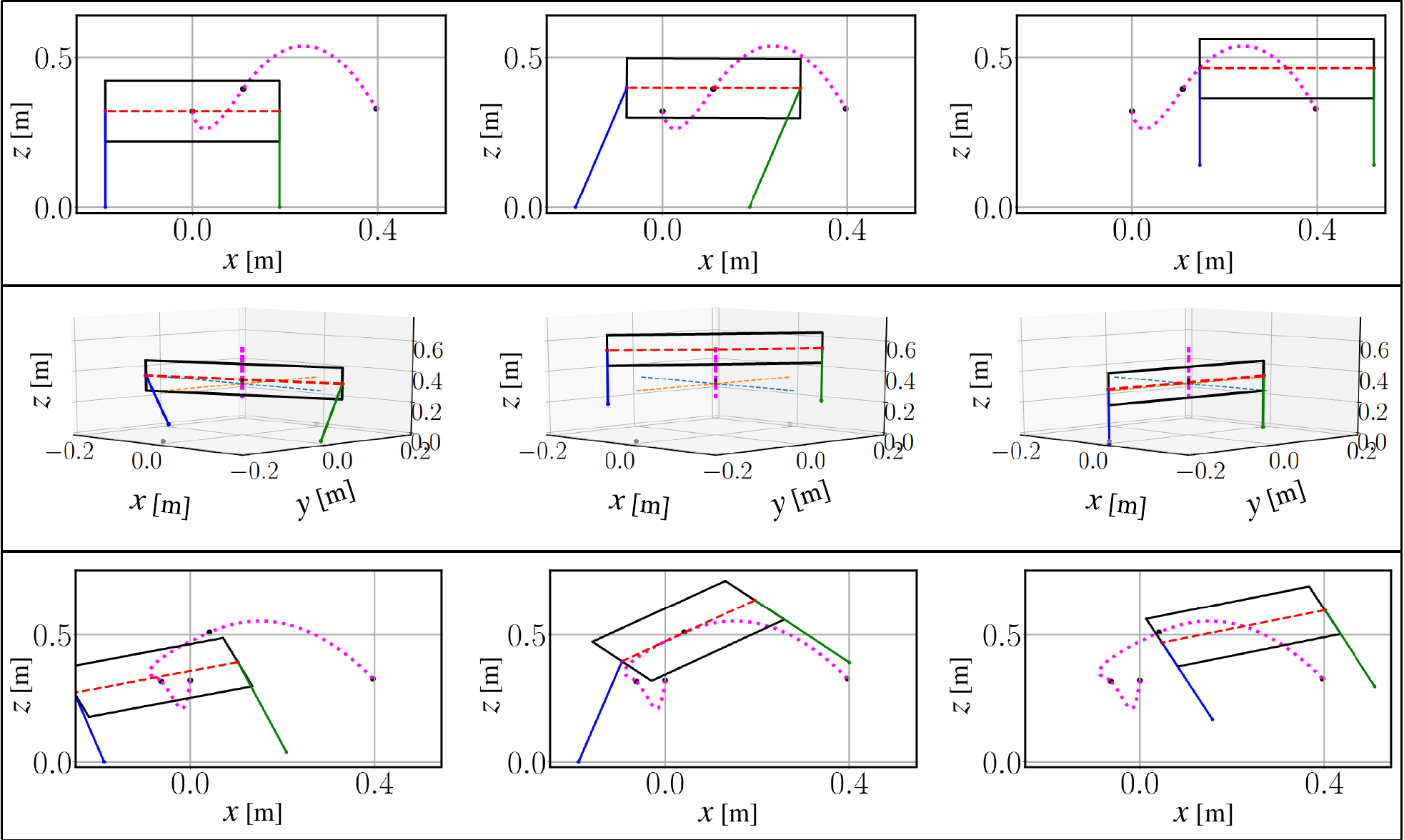}
			\caption{Versatile explosive motions generated by the dual-layer TO. From top to bottom, the robot performs: 1) $0.4\,$m forward pronking, 2) jumping with $90^{\circ}$ spinning, and 3) $0.4\,$m froggy jumping. The blue and green solid lines separately represent the rear and front legs, while the black box marks the trunk. The red dashed lines attached to the trunk represent the body rotation. The pink dotted curves in each figure represent CoM trajectories, while the black dots on them mark the way-points. Detailed motions can be found in the supplementary video.}
			\label{fig:quadrupedal_motion_simu}
		\end{figure}
		
		\begin{figure}
			\centering
			\includegraphics[width=\columnwidth]{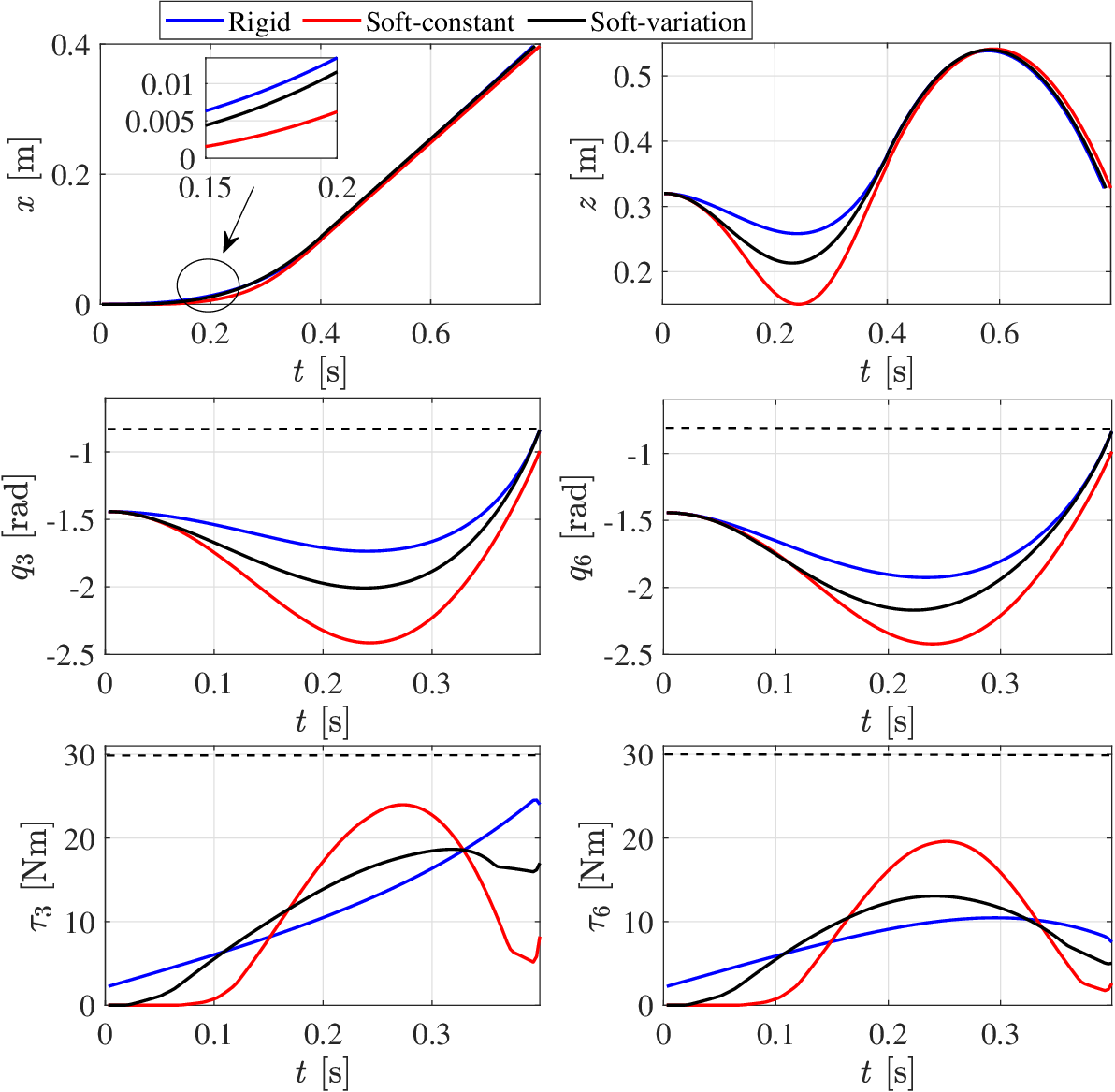}
			\caption{$40\,$cm forward pronking for rigid and articulated robots. `Rigid', `Soft-constant' and `Soft-variation' separately denote the motion generated by the TD-aSLIP model without parallel compliance, with constant stiffness and with varying stiffness. From the top to bottom, the sagittal CoM trajectory, calf joint angles and calf joint torques are plotted. $q_3$ and $q_6$ separately denote the front calf angle and rear calf angle, $\tau_3$ and $\tau_6$ separately denote the front calf torque and rear calf torque. Here, we set the rest length $l^1_0$ = $l^2_0$ = $0.32\,$m. The dashed lines in the second and third rows mark the upper boundaries.}
			\label{fig:quadrupedal_motion_rigid_compliance}
		\end{figure}

		\subsection{Explosive motion with compliance} \label{motion_with_varying_stiffness} 
		The section investigates the benefits of parallel compliance in achieving explosive motion.
		As detailed in \eqref{equ:assp1_slip2}, both the rest length (i.e., $l^1_0$ and $l^2_0$) and the spring stiffness (i.e., $k_s^1$ and $k^2_s$) affect the performance.
		\subsubsection{Motion generation with/without parallel compliance}
		To start, we generate the forward pronking motion using the TD-aSLIP model with different compliance settings. Fig.~\ref{fig:quadrupedal_motion_rigid_compliance} compares the pronking movements of the rigid robot ($k_s^1$ = $k_s^2$ = $0\,$N/m) and the articulated soft robot with constant spring stiffness ($k_s^1$ = $k_s^2$ = $1000\,$N/m\footnote{The equivalent stiffness of parallel springs is computed when the robot stands in the homing pose (shown in Fig.~\ref{fig:quadrupedal_slip_model}(b)). The resultant stiffness is marked by the red star in Fig.~\ref{fig:stiffness_mapping}. Here we double it for the SLIP model.}). Furthermore, the pronking motion for the compliant robot with varying stiffness (computed by Section~\hyperlink{sec:nonlinear_stiffness}{IV-B2}) is also plotted in Fig.~\ref{fig:quadrupedal_motion_rigid_compliance}.
		
		As can be seen in Fig.~\ref{fig:quadrupedal_motion_rigid_compliance}, the proposed dual-layer TO successfully obtained optimal motions, including feasible Cartesian trajectories (see the first row), joint movements (see the second row) and torque profiles (see the third row), for three cases. With compliance, lower height in the stance phase is achieved, among which the TD-aSLIP with constant spring stiffness (denoted as `soft-constant') contributes to the lowest height (see $z$ plots before $t$ = $0.4\,$s in the first row of Fig.~\ref{fig:quadrupedal_motion_rigid_compliance}). Statistics in Table~\ref{table:energy_performance_x} reveal that a lower peak torque (`$\tau^{\max}$') and a lower peak power (`$P^{\max}$') is needed when parallel compliance exists. Compared with the rigid case, more total mechanical energy (`$E$-tot') is needed since extra Cartesian motion, e.g., vertical motion, is required in the stance phase. However, the energy needed by the motor actuation (`$E$-actu') is reduced, meaning that the springs contribute to an energy-saving motion in the feedforward control manner.
		\begin{table}
			\centering
			\caption{{Energetic performance of different models. Here, we compute the mechanical power and energy.}}
			\label{table:energy_performance_x}
			\begin{tabular}{c|c|c|c}
				\toprule
				{}&{`Rigid'}&{`Soft-constant'}&{`Soft-variation'}\\
				\hline
				{{$\tau^{\max}$ [\text{Nm}] }}&{24.5}&{22.1}&{20.2}\\
				\hline
				{{$P^{\max}$ [\text{W}] }}&{1253.0}&{590.9}&{924.8}\\ 
				\hline
				{{$E$-tot [J]}}&{69.0}&{76.2}&{74.3}\\	
				\hline
				{{$E$-actu [J]}}&{69.0}&{19.3}&{28.5}\\	           
				\bottomrule  
			\end{tabular}
		\end{table}

		\subsubsection{Pronking motion with different spring settings}\label{pronking_performance_spring_setting}
		To investigate the effectiveness of parallel compliance further, we generate pronking motions with different spring settings by changing the spring stiffness and the rest length. Fig.~\ref{fig:quadrupedal_motion_compliance_variation} shows the explosive pronking motions, for brevity, in the stance phase and Fig.~\ref{fig:pronking_eneergy_compliance_variation} reports the energetic performance. 
		
		The first row in Fig.~\ref{fig:quadrupedal_motion_compliance_variation} compares the sagittal movement (in `$x\!-\!z$' plane) regarding stiffness variation, where the equivalent stiffness is indicated by the `homing stiffness' such as `0' and `250'. As it can be seen, as stiffness increases, the minimal vertical height reduces. Furthermore, more forward (`$x$') movement is needed (see the partially enlarged drawing on the top left) as the stiffness increases. As a result, more total energy (see `$E-$tot' with `$(l_0^1,l_0^2)$=0.32' in Fig.~\ref{fig:pronking_eneergy_compliance_variation}) is required. However, smaller peak power (see `$P^{\max}$' with `$(l_0^1,l_0^2)$=0.32' in Fig.~\ref{fig:pronking_eneergy_compliance_variation}) is required to accomplish this jumping task. In addition, smaller peak torque (see `$\tau^{\max}$' with `$(l_0^1,l_0^2)$=0.32' in Fig.~\ref{fig:pronking_eneergy_compliance_variation}) is required when parallel compliance is exploited, and $k^1_s$ = $k^2$ = $1000\,$N/m results in the lowest $\tau^{\max}$. 
		
		With $k_s^1$ = $k_s^2$ = $1000\,$N/m, the second row in Fig.~\ref{fig:quadrupedal_motion_compliance_variation} compares the `$x\!-\!z$' movement regarding different rest lengths. In particular, we present the results with the rest length being $0.32\,$m, $0.38\,$m and $0.42\,$m, where $0.38\,$m and $0.42\,$m are separately computed by summing the two link lengths with/without considering the physical joint limits. It turns out that, as the rest length increases, the minimal height increases (see the `$z$' curves on the bottom right) while the forward motions are almost the same (see the `$x$' curves on the bottom left). As illustrated by Fig.~\ref{fig:pronking_eneergy_compliance_variation}, a larger rest length results in a larger peak torque `$\tau^{\max}$', a larger peak power `$P^{\max}$' but a smaller energy cost `$E$-tot'. 
		
		Furthermore, Fig.~\ref{fig:pronking_eneergy_compliance_variation} demonstrates that, no matter which rest length is used, the robot with parallel compliance (when the `homing stiffness' $>$ 0 ) needs a smaller peak torque and a smaller peak power. To achieve the best performance within the kinematic limit, we then set $k^1_s$ = $k^2_s$ = $1000\,$N/m and $l_1^0$ = $l_2^0$ = $0.38\,$m for the hardware tests if no exception is mentioned. The reason is that this combination of parameters achieves the least `$\tau^{\max}$' without increasing `$E$-tot too much. Also, this equivalent spring stiffness can be easily achieved by attaching the standard springs to the hardware. 
		
		\begin{figure}
			\centering
			\includegraphics[width=\columnwidth]{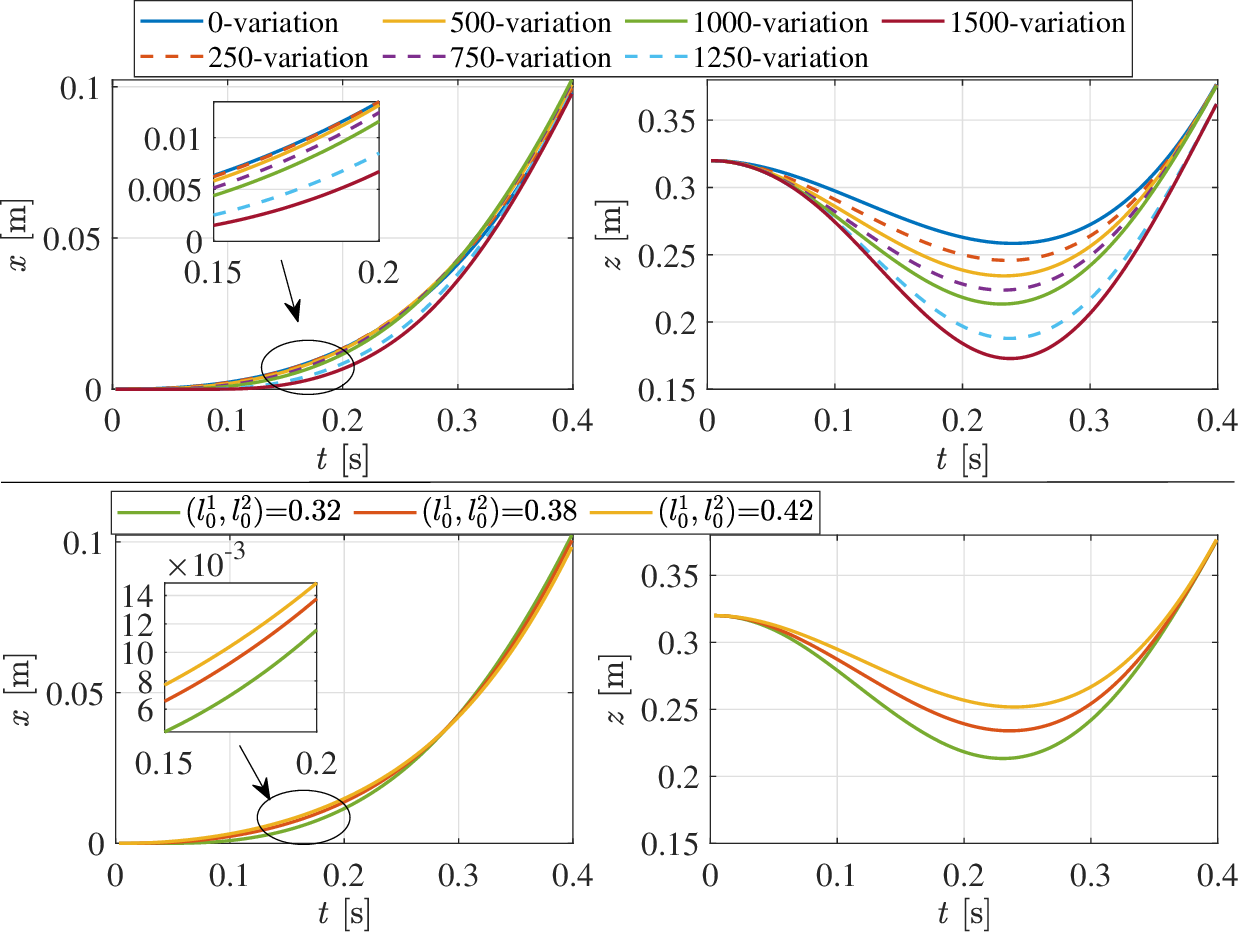}
			\caption{Sagittal pronking motion with different spring settings. The first row demonstrates the $x\!-\!z$ motion with different spring stiffness, of which the `homing stiffness' is indicated by the legends. In this case, $l^1_0$ = $l^2_0$ = $0.32\,$m. The second row displays the motion with different rest lengths, as indicated by the legends. In this case, $k^1_s$ = $k^2_s$ = $1000\,$N/m. Note that the varying stiffness mapping is used in all cases.}
			\label{fig:quadrupedal_motion_compliance_variation}
		\end{figure}
		
		\begin{figure}
			\centering
			\includegraphics[width=\columnwidth]{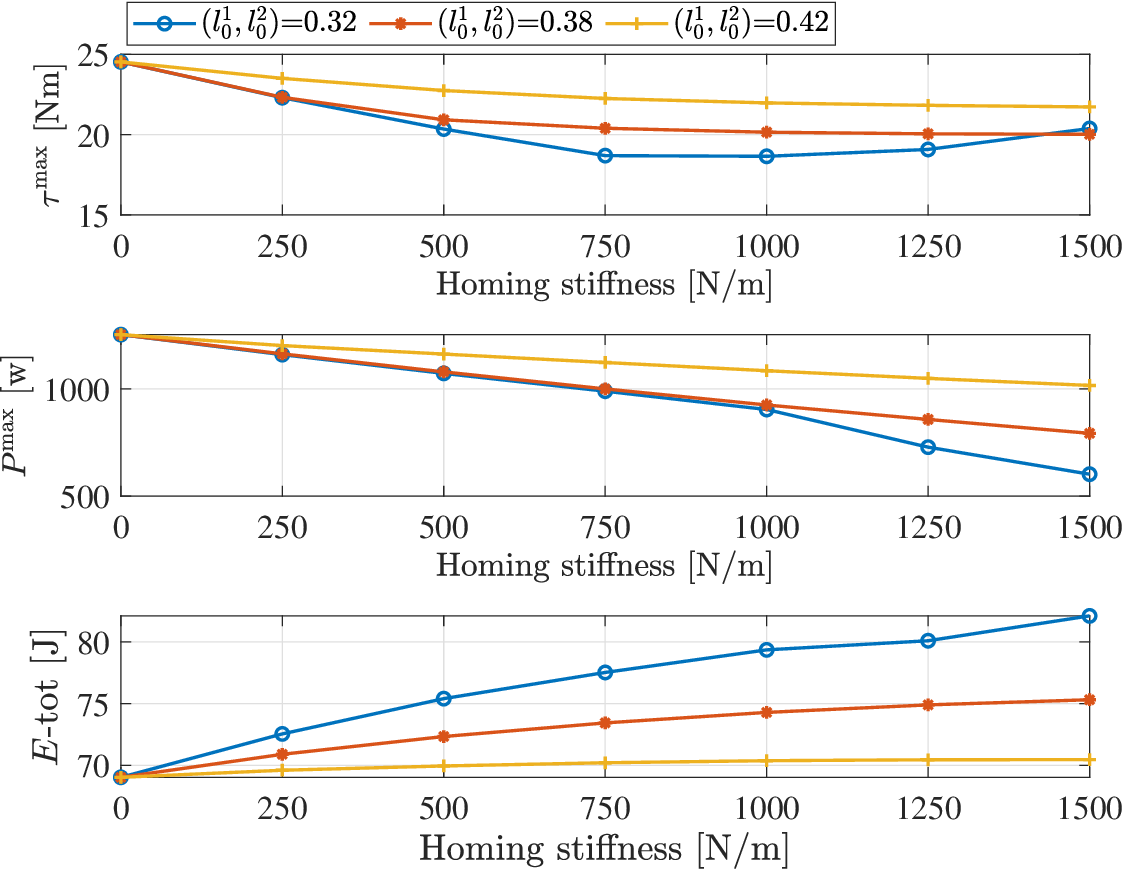}
			\caption{Pronking performances with different compliance settings. The first, second and third rows separately demonstrate the peak torque, the peak mechanical power and the total mechanical energy cost. Note that `$(l_0^1,l_0^2)=0.38$' is approaching the joint limit. We also present the result with `$(l_0^1,l_0^2)=0.42$', which goes beyond the physical limit, to study the general rules.}
			\label{fig:pronking_eneergy_compliance_variation}
		\end{figure}

		\subsection{TD-aSLIP vs dual-aSLIP} \label{sec:jumping_motion}
		Compared to traditional SLIP models, such as the dual-aSLIP model in (\cite{ding2024Quadrupedal}), the proposed TD-aSLIP model is able to characterize explosive motion requiring large body motions. For explosive motion without requiring large body rotation, we here compare the tracking performance using two models.
		Taking the rigid case, for example, we compare the tracking performance of pronking motion under the reference trajectories generated by the TD-aSLIP model and the dual-aSLIP model. The baseline motion without utilizing the trunking rotation was generated by the similar dual-layer TO scheme except that the body rotation is not optimized. In both cases, the quadruped was controlled with the torque controller in Section~\hyperlink{sec_compliance_control}{VI}, sharing the same control gains. The dynamic simulation is conducted with \texttt{PyBullet} simulator (\cite{coumans2016pybullet}). 
		
		The forward CoM trajectories and pitch angle profiles of the $40\,$cm pronking jumping are plotted in Fig.~\ref{fig:pronking_comparison}. 
		\begin{figure}
			\centering
			\includegraphics[width=\columnwidth]{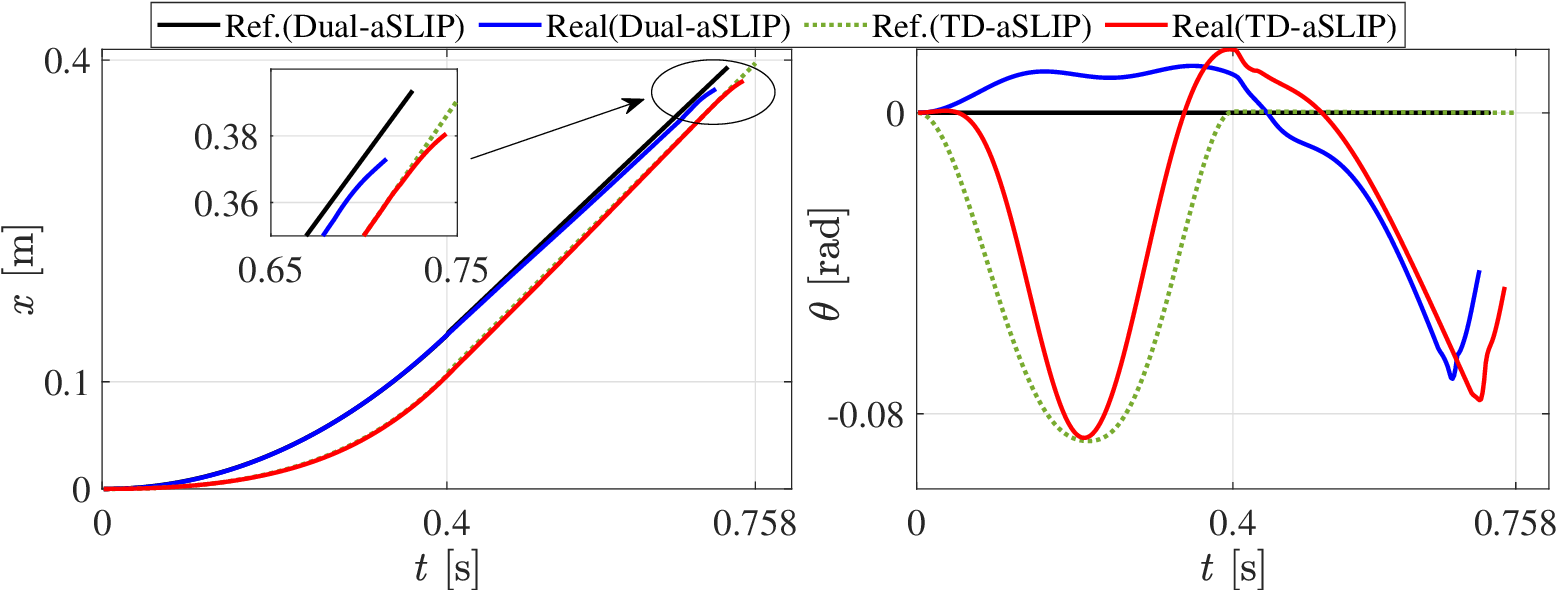}
			\caption{$40\,$cm pronking motion with different SLIP models. `Ref.' and `Real' separately denote the reference and measured trajectories. Note that the durations of two reference motions are different from each other.}
			\label{fig:pronking_comparison}
		\end{figure}
		As can be seen from the right side of Fig.~\ref{fig:pronking_comparison}, although the reference pitches passing the waypoints are set to be zeros, an optimal pitch trajectory is generated with the TD-aSLIP model (see the green dotted `Ref.(TD-aSLIP)'). That is, the body rotation is utilized for pronking with the TD-aSLIP model. As a result, a longer final CoM position is achieved (see the partially enlarged drawing on the left side of Fig.~\ref{fig:pronking_comparison}) and the smaller final CoM trajectory error is obtained (see the `CoM error' with `Desired Distance'=$0.4\,$m on the bottom row of Fig.~\ref{fig:pronking_tracking_error_comparison}). Furthermore, as can be seen from the top row of Fig.~\ref{fig:pronking_tracking_error_comparison},
		the TD-aSLIP also results in a smaller error in the landing position. That is, the TD-aSLIP is more accurate, enhancing tracking performance. 
		
		In addition, Fig.~\ref{fig:pronking_tracking_error_comparison} also reports the tracking errors when the desired jumping distances range from $0.0\,$m to $0.6\,$m. It turns out that, compared to dual-aSLIP, the TD-aSLIP achieves a smaller landing error in most cases, e.g., from $0.2\,$m to $0.6\,$m. Furthermore, in all tests, better CoM tracking is achieved with the TD-aSLIP.  That is, the proposed model exploits body rotation to enhance explosive pronking.
		\begin{figure}
			\centering
			\includegraphics[width=\columnwidth]{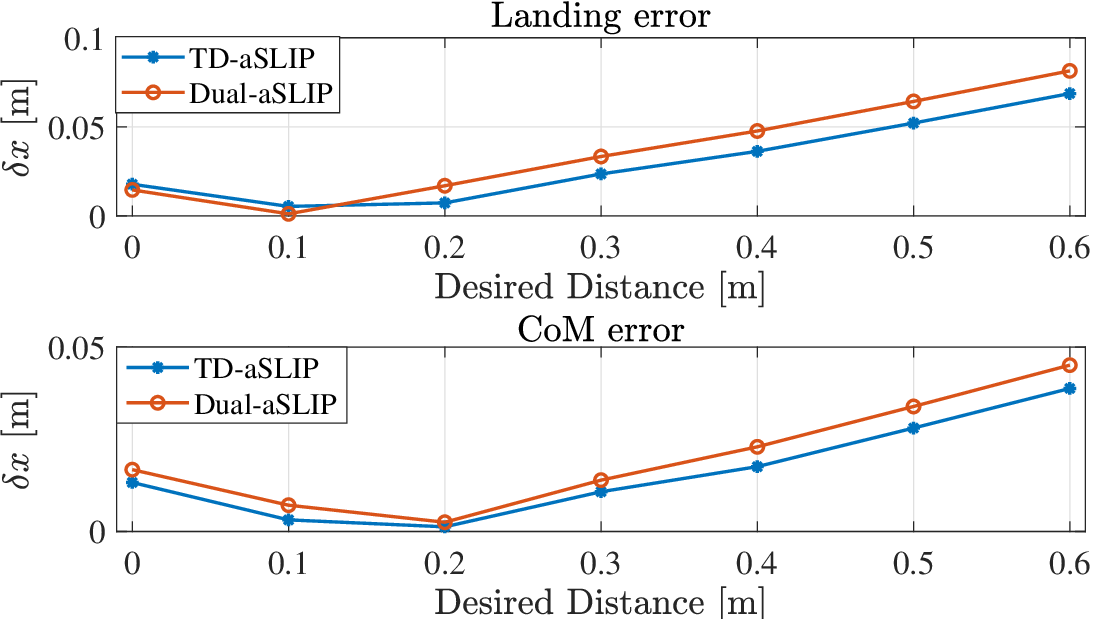}
			\caption{Forward tracking errors (`$\delta x$') when jumping with SLIP models. `Landing error' refers to the tracking error of the forward landing position and `CoM error' refers to the tracking error of the forward CoM position when detecting landing.}
			\label{fig:pronking_tracking_error_comparison}
		\end{figure}



		\section{Hardware Experiments} \label{sec_hardware_evaluation}
		The section presents the experimental results. To start, we introduce the newly designed PEA-driven quadruped, i.e., E-Go-V2. Then, we validate the versatile, robust, and explosive motion with hardware experiments. In particular, we highlight the benefits of parallel compliance with comparison studies.
		
		\begin{figure}
			\centering
			\includegraphics[width=\columnwidth]{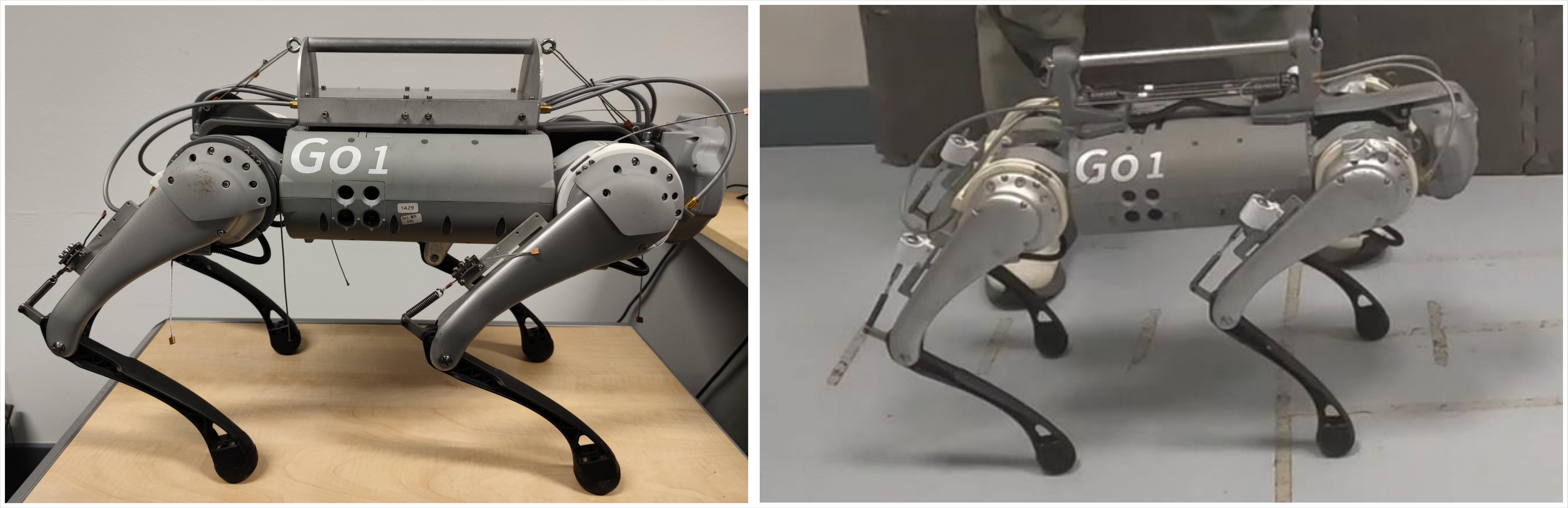}
			\caption{The articulated compliant quadruped with the parallel springs: from Delft E-Go (left) to E-Go-V2 (right) 
				.}
			\label{fig:egov2_VS_egov1}
		\end{figure}

		\begin{figure}
			\centering
			\includegraphics[width=\columnwidth]{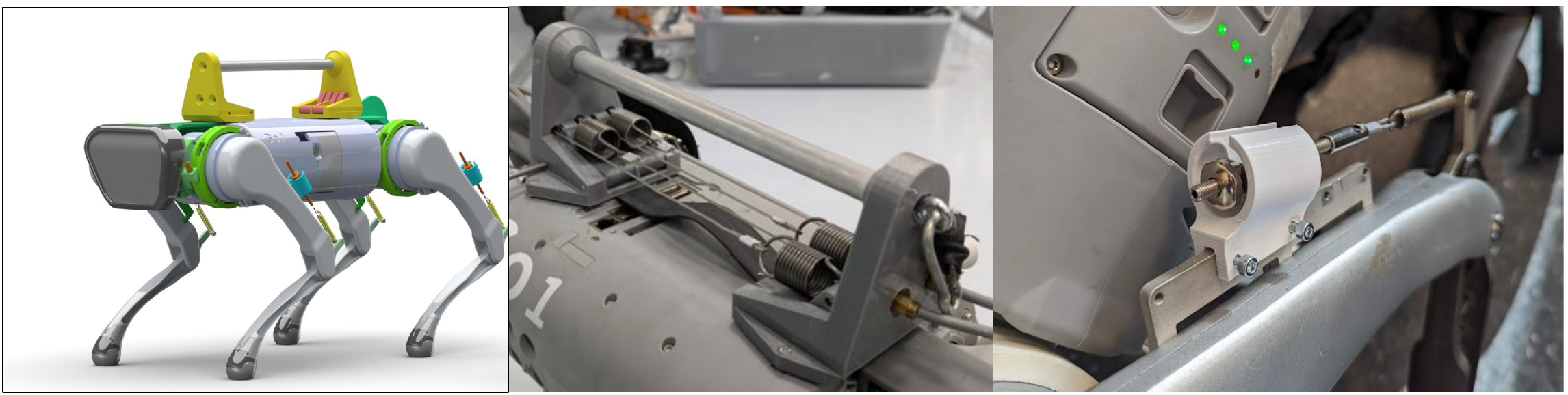}
			\caption{E-Go-V2 design. The sub-figures from left to right separately illustrate the 3D rendering, the engagement mechanism for thigh springs and the engagement mechanism for calf springs.}
			\label{fig:egov2_spring}
		\end{figure}
		
		\begin{figure}
			\centering
			\includegraphics[width=\columnwidth]{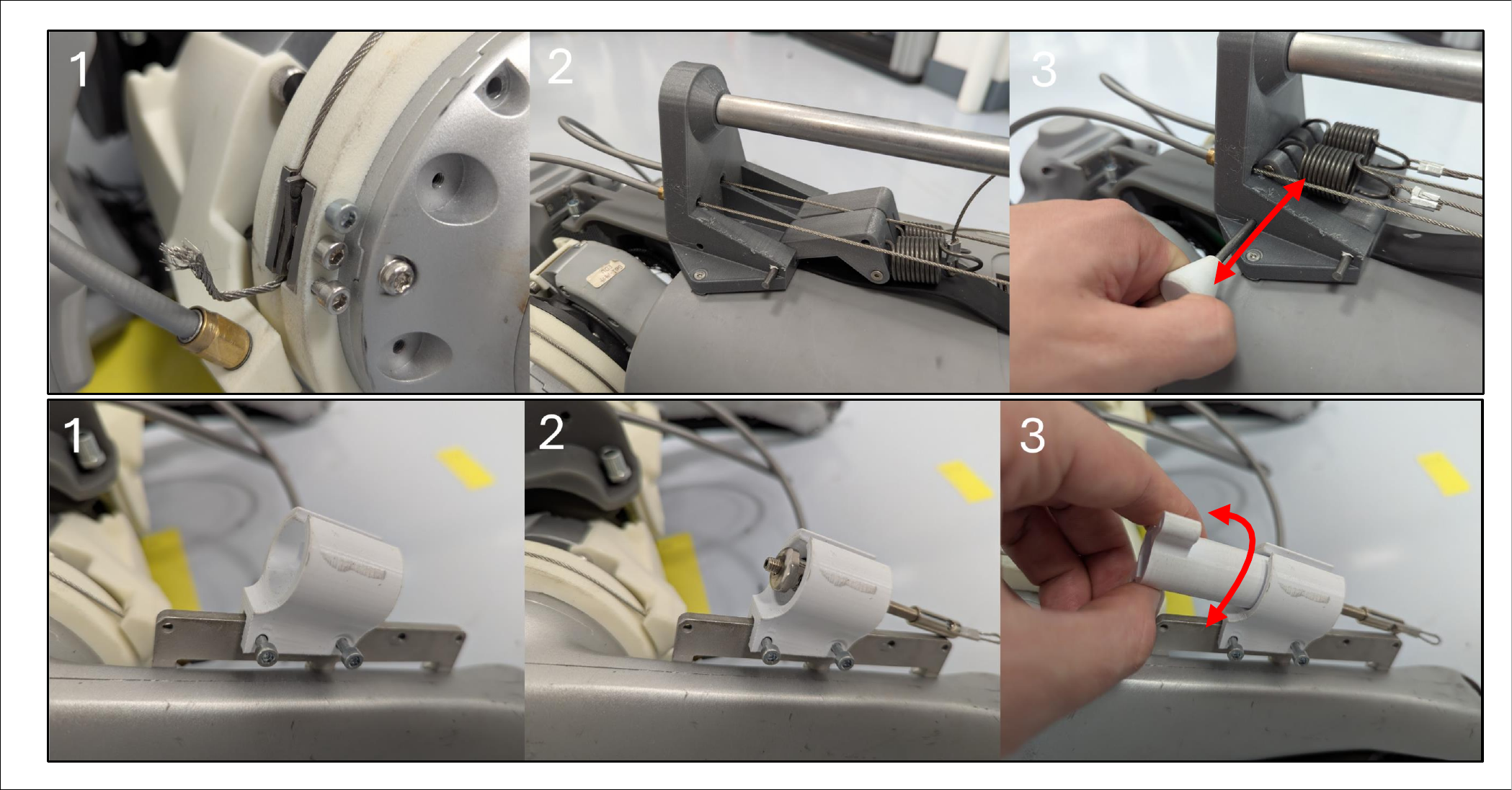}
			\caption{The engagement process for thigh spring (top) and calf spring (bottom). On the top, step 1 sets the spring's rest length, while 2 and 3 separately denote the release and engagement configurations. On the bottom, 1 and 2 separately denote the release and engagement configurations, while step 3 adjusts the rest length.}
			\label{fig:egov2_spring_engagement}
		\end{figure}
		\subsection{Experimental setup}
		\subsubsection{Hardware design}
		An easily-accessible articulated soft quadruped plays a crucial role in validating our study. Instead of manufacturing a robot from scratch, we design the PEA-driven quadruped robot by enhancing the commercially available Unitree Go1 with spring add-ons, see the right picture in Fig.~\ref{fig:egov2_VS_egov1}. In this new platform, parallel tension springs are attached to sagittal joints, including the thigh joint and calf joint on each leg. Differing from the previous version introduced in (\cite{ding2024Quadrupedal}) (see the left picture in Fig.~\ref{fig:egov2_VS_egov1}), we improve the design by 1) removing the hip springs to make the whole system more compact and lightweight, 2) redesigning the engagement mechanism (see the middle picture in Fig.~\ref{fig:egov2_spring}) for the thigh springs to reduce friction, 3) redesigning the engagement mechanism (see the right picture in Fig.~\ref{fig:egov2_spring}) of calf springs to make the connection more solid. Furthermore, the novel platform enables us to engage/disengage springs quicker and reset the rest length more easily. The user can also replace the springs if needed. The engagement operation, together with the adjustment of rest length, is illustrated in \ref{fig:egov2_spring_engagement}. {To prompt the research in the field of compliant quadrupeds, we open-source the CAD files of the hardware design. Please check \url{https://github.com/jtdingx/Delft-E-Go-quadruped/tree/E-go-V2}}

		\begin{figure}
			\centering
			\includegraphics[width=\columnwidth-2mm]{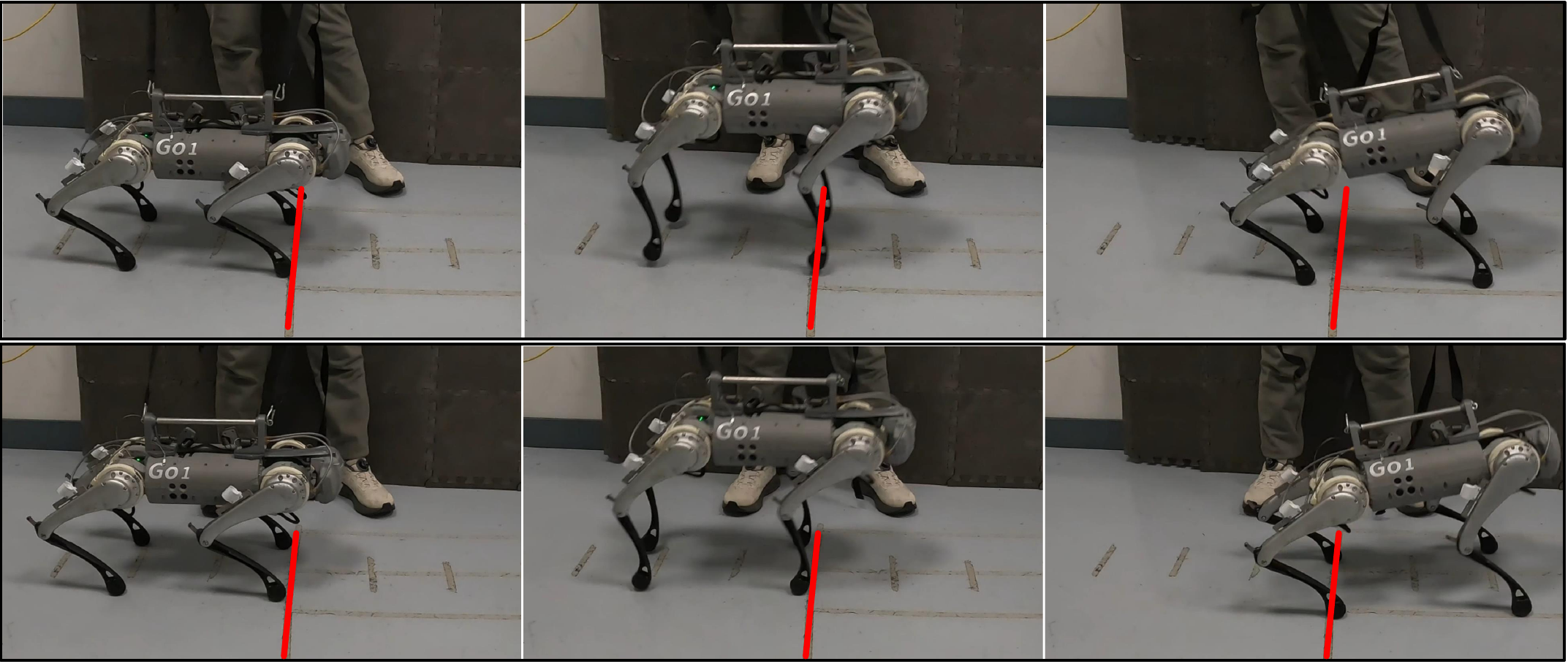}
			\caption{$40\,$cm forward pronking motions under different template models. The top panel demonstrates the motion achieved with the dual-aSLIP model without considering body rotation, while the bottom panel is achieved with the TD-aSLIP model. In each snapshot, the red solid line marks the desired landing position for the rear legs.}
			\label{fig:enhance_pronking_novel_slip}
		\end{figure}
		
		\subsubsection{System setting}
		All the hardware experiments are conducted with E-Go-V2. Without exceptional explanation, the spring stiffness of the thigh and calf springs are $11.4\,$N/m and $23.5\,$N/m, respectively, and the equivalent stiffness for the TD-aSLIP model at the homing pose is about $1000\,$N/m. 
		
		This work focuses on explosive motion with a flight phase when all legs lift off. To achieve a stable landing, we estimate the duration of the real flight phase and then set the real body inclination (including roll and pitch) as the reference after the middle flight. Upon landing, the reference horizontal CoM position immediately changes to the estimated landing position, the reference height is set to the sum of the homing height and the estimated foot landing height, and the reference velocity is set to zero. At the same time, an exponential function is used to generate the reference body inclination angle, achieving a smooth body rotation. An example is provided in Fig.~\ref{fig:40cm_forward_pronking_plot}, see the yellow curves.
		
		For robot control, we use a Kalman filter inspired by (\cite{bledt2018cheetah}) to estimate the state. Since the IMU sensory data drifted a lot in the explosive motion process, we attached the landmarks to the ground to measure the real jumping distance. One example is shown in Fig.~\ref{fig:enhance_pronking_novel_slip}. The distance between the neighboring gray strips is $20\,$cm. For tracking control, the code runs in \texttt{c++}. The MPC and WBC are solved via \texttt{OSQP} (\cite{osqp}). In addition, it is worth mentioning that all locomotion control tasks share the same control gains.
		
		\subsection{Versatile \& Explosive motion- enhanced dynamic behavior}
		With the above platform, we validate the versatile, explosive motion in this section. In particular, we highlight the enhanced dynamic behavior by attaching parallel springs. 
		
		\subsubsection{Enhanced pronking} 
		We start from the pronking motion, where a large body rotation is not required. First, we compare the explove pronking with different template models. Then, we investigate the pronking with parallel compliance. 
		
		\emph{TD-aSLIP vs dual-aSLIP}: Fig.~\ref{fig:enhance_pronking_novel_slip} illustrates the $40\,$cm forward pronking with different SLIP models, where the top row shows the motion following the reference generated by dual-aSLIP while the bottom row shows the motion with TD-aSLIP. As it can be seen, the TD-aSLIP model enables the robot to land at the desired location (marked by the red line) while the dual-aSLIP landed behind. The plots in Fig.~\ref{fig:40cm_forward_pronking_plot} show that, since the body rotation is utilized (see the `Ref.(TD-aSLIP)' in the `$\theta-t$' plot), the robot achieves a higher peak height (see the `$z-t$' plot) and a longer flight phase (see the partially enlarged drawing in the `$x-t$' plot). As a result, the robot reached the desired distance\footnote{The yellow curve in the partially enlarged drawing (in the `$x-t$' plot) shows that the estimated landing position fell a little behind the desired distance, i.e., $40\,$cm. However, the robot in the second row of Fig.~\ref{fig:enhance_pronking_novel_slip} indeed reached $40\,$cm, as indicated by the land marker.}.

		\begin{figure}
			\centering
			\includegraphics[width=\columnwidth-0mm]{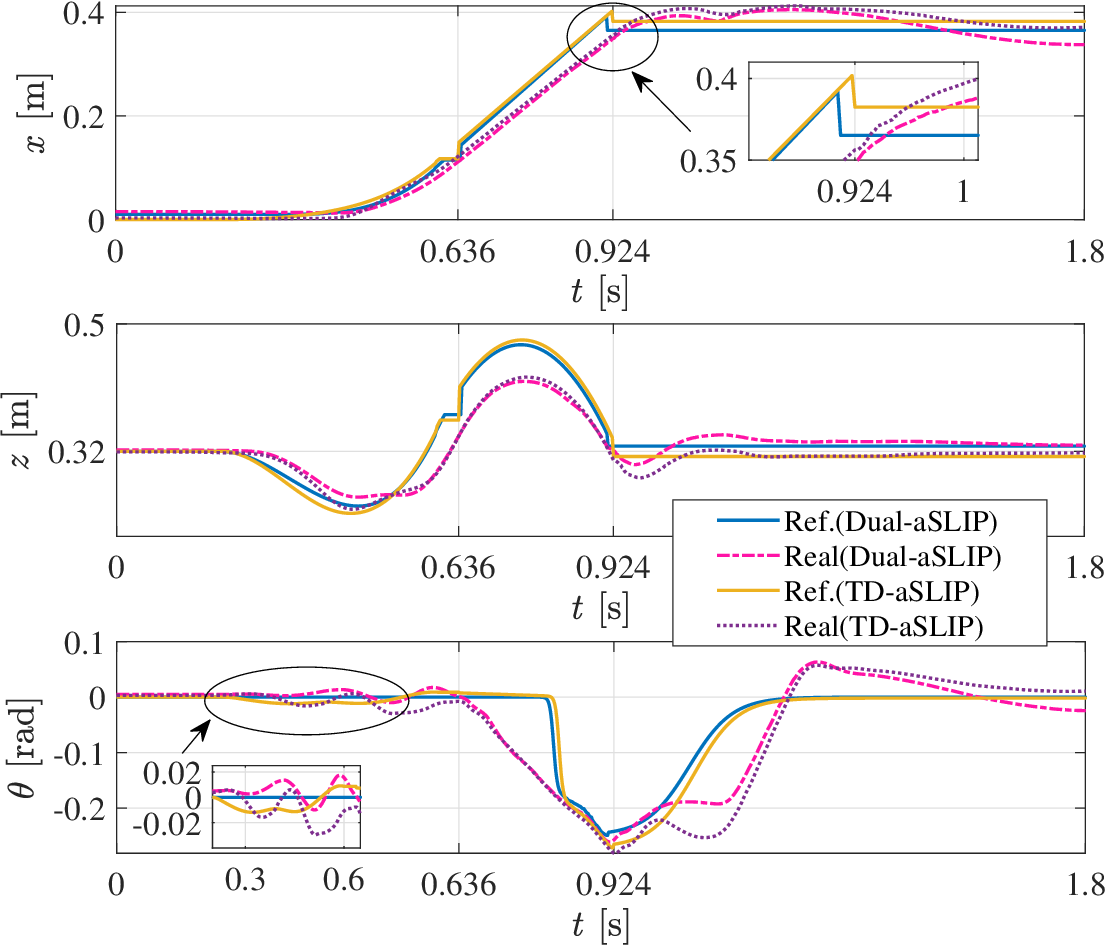}
			\caption{Trajectory profiles for $40\,$cm forward pronking with the dual-aSLIP model and TD-aSLIP model. `0.636' and `0.924' separately mark the take-off and landing moments when using the TD-aSLIP model. Since the robot reached the kinematic limits before entering the flight phase, e.g, $0.636\,$s with the TD-aSLIP model, the robot tracked a constant reference that keeps unchanged in the later stance phase until all the legs lifted off.}
			\label{fig:40cm_forward_pronking_plot}
		\end{figure}

		\emph{Pronking with parallel compliance}: With springs engaged, the robot can also achieve $40\,$cm forward pronking. The snapshot of one trial is illustrated in the first row of Fig.~\ref{fig:snaptshots}. Furthermore, with the TD-aSLIP model, the compliant robot with springs engaged can achieve $50\,$cm pronking. The comparison motions between the rigid robot and the compliant robot are presented in the first group of Fig.~\ref{fig:enhance_tracking_performance}, of which the trajectory profiles are displayed in Fig.~\ref{fig:50cm_forward_pronking_springs_plot}.
		
		Fig.~\ref{fig:50cm_forward_pronking_springs_plot} demonstrates that, when parallel compliance is considered, the reference motion changes. The longer forward position between $0.626\,$s and $0.918\,$s (comparing the blue curve and the yellow curve in the `$x-t$' plots) indicates that the robot with parallel compliance (denoted as `(Soft)') would accomplish the pronking motion within a shorter period, meaning a more aggressive jumping. Nevertheless, due to the existence of parallel springs, the compliant robot required a smaller peak torque than the rigid robot, as can be seen by observing the `$\tau$-calf' plots from $0.3\,$s to $0.626\,$s at the bottom of Fig.~\ref{fig:50cm_forward_pronking_springs_plot}. In this case, the rigid robot also suffered from a large pitch movement in the fight phase, as can be seen by comparing the `Real(Rigid)' and `Real(Soft)' between $0.626\,$s and $0.918\,$s in the `$\theta-t$' plot. As a result, the compliant robot landed at the desired location while the rigid robot only landed at about $40\,$cm, see the fifth column (marked by the purple bracket) in the first group of Fig.~\ref{fig:enhance_tracking_performance}.
		
		Furthermore, the torque profiles, i.e., `$\tau$-thigh' and `$\tau$-calf' plots, in Fig.~\ref{fig:50cm_forward_pronking_springs_plot} show that the rigid robot required a larger torque after landing (after $0.918\,$s) so as to return to the homing pose. In contrast,allel compliance reduces the torque burden after landing.
		
		The `$x-t$' plots in the top row show that the reference forward CoM positions did not change for a while before $0.626\,$s in both the rigid and soft cases. This is because the reference approached the joint limits and the later take-off happened. Thus, we kept the reference unchanged (near the feasible bound) until the robot took off. A similar phenomenon also occurs on the reference `$\theta$' trajectory.
		
		\begin{figure}
			\centering
			\includegraphics[width=\columnwidth-0mm]{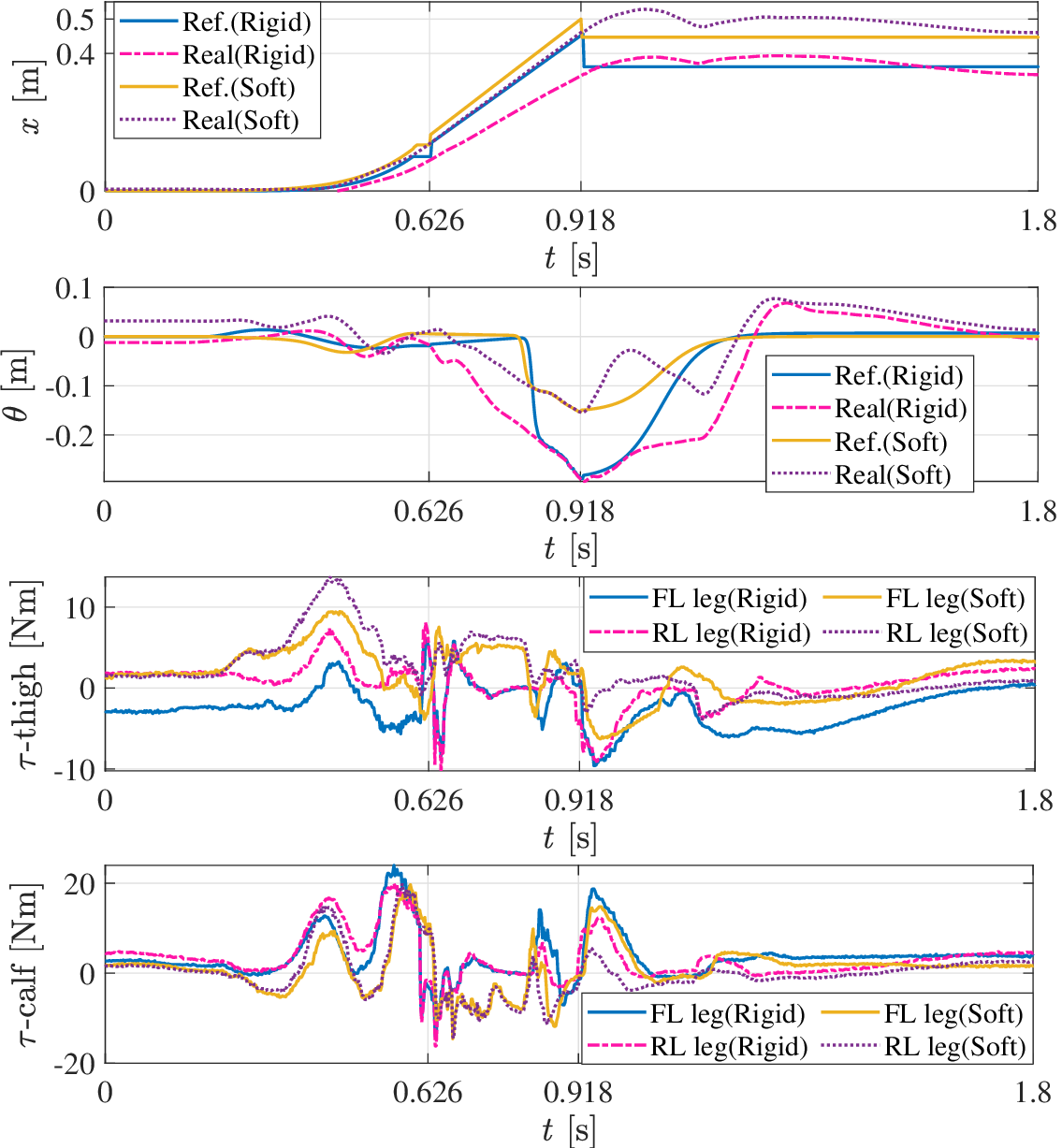}
			\caption{Trajectory profiles for $50\,$cm forward pronking with the rigid and soft quadrupeds. `0.626' and `0.918' separately denote the take-off and landing moments when parallel springs are engaged.}
			\label{fig:50cm_forward_pronking_springs_plot}
		\end{figure}
		\begin{figure*}
			\centering
			\includegraphics[width=2\columnwidth+2mm]{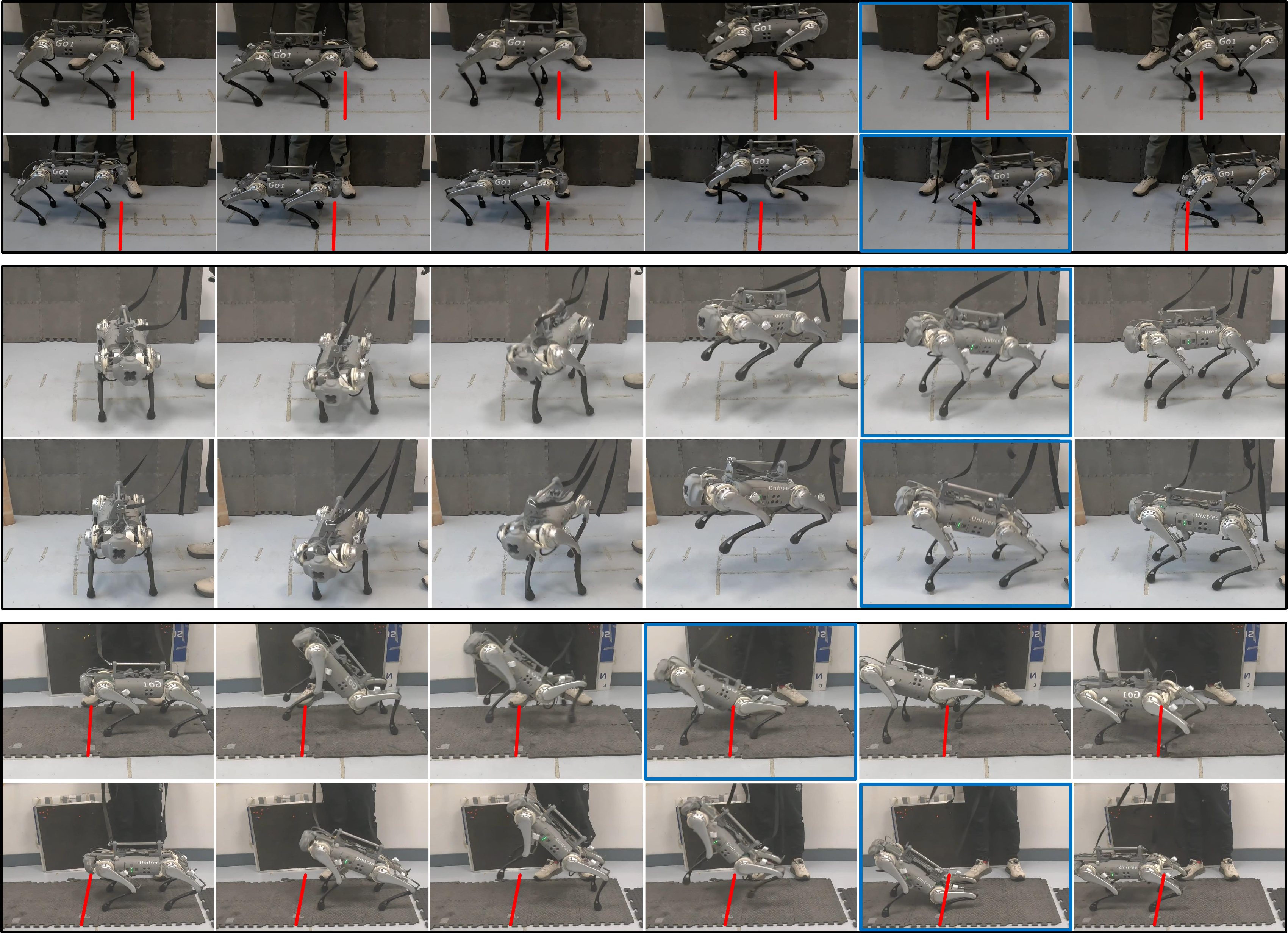}
			\caption{Comparative explosive motion between the robot with/without parallel springs engaged. From top to bottom, we present the comparison motions for $50\,$cm forward pronking (top group), $90^\circ$ clockwise hop-turn  (middle group), $50\,$cm froggy jumping (bottom group). In each group, the upper panel demonstrates the motion with a rigid robot while the bottom demonstrates the compliant case with springs engaged. The red lines (if appearing) mark the desired landing positions for the rear legs.}
			\label{fig:enhance_tracking_performance}
		\end{figure*}

		\subsubsection{Enhanced hop-turn}\label{enhance_hop_turn}
		Hop-turn is one typical motion requiring a large yaw motion. The second group in Fig.~\ref{fig:50cm_forward_pronking_springs_plot} compares the $90^\circ$ clockwise yaw-jumping with/without springs engaged. It turns out that the rigid robot could only land with about $70^\circ$ yaw while the compliant robot landed with about $85^\circ$ yaw (comparing the fifth column of the second group in Fig.~\ref{fig:50cm_forward_pronking_springs_plot}). That is, the incorporation of parallel compliance improves the tracking performance. Further experiments reveal that the compliant robot with springs engaged can realize $145^\circ$ hop-turn at most, whereas the rigid case only achieves $125^\circ$ hop-turn.
		
		It is worth mentioning that our tracking controller can also realize a $90^\circ$ hop-turn if we reduce the homing height, e.g., from $0.32\,$m to $0.25\,$m or $0.2\,$m. Detailed motions are attached in the supplementary video. The low homing height is also evidenced by many existing works such as (\cite{song2022optimal}) and (\cite{garcia2021time}). Here, for comparison studies, we still keep the homing pose to the normal value in this section.

		\subsubsection{Enhanced froggy jumping}
		As an explosive motion requiring a large body rotation, froggy jumping is also compared. Note that we here reduce the homing height from $0.32\,$m to $0.25\,$m to obey kinematic limits, especially those involving the rear legs. It turns out that both the rigid and compliant robots can achieve $40\,$cm forward froggy jumping. One of the $40\,$cm froggy jumping with the compliant robot is demonstrated in the second row of Fig.~\ref{fig:snaptshots}.
		
		Moving further, we investigate that extreme froggy jumping with/without parallel compliance. Experiments demonstrate that the compliant robot could realize at most $50\,$cm forward froggy jumping while the rigid robot can not. One comparison is displayed in the third group in Fig.~\ref{fig:enhance_tracking_performance}. Fig.~\ref{fig:50cm_froggy_springs_plot} compares the motion and torque profiles for both rigid and compliant cases. 
		
		As marked in Fig.~\ref{fig:50cm_froggy_springs_plot}, for the compliant robot, the half flight phase (front legs lift off), flight phase (all the legs lift off) and the landing phase (at least two legs touch down) separately started at $0.266\,$s, $0.532\,$s, and $0.802\,$s. Compared to the rigid case, the compliant robot landed later, accompanied by a larger landing distance. In both cases, before entering the flight phase, the largest torque occurred in the FL calf joint with the rigid robot, which is about $24.3\,$Nm right after $0.266\,$s. After taking off, the rigid robot required the largest torque, which is around $-30\,$Nm right after $0.532\,$s (see the `RL leg(Rigid)' curve in the `$\tau$-calf' plots).
		
		It should be mentioned that upon entering the flight phase (after $0.532\,$s), the robot needs to keep the rear legs contract to avoid collision with the ground. When parallel springs are engaged, the compliant robot needs to output a larger torque to overcome the unidirectional spring force, as illustrated by the `RL leg(Soft)' curve in the `$\tau$-thigh' plot. Since the rigid robot landed early with a larger forward velocity, it kept moving by rotating around the front contact line. To maintain balance, the FL leg outputs a large thigh torque, as can be seen from the `FL leg(Rigid)' curve in the `$\tau$-thigh' plot around $0.8\,$s. The post-landing motion of the rigid robot is demonstrated by the fifth and sixth columns in the fifth row of Fig.~\ref{fig:enhance_tracking_performance}.
		
		\begin{table}
			\centering
			\caption{Comparison of extreme explosive motion with/without parallel compliance.}
			\label{table:enhanced_tracking_percentage}
			\begin{tabular}{c|c|c|c}
				\toprule
				{}&{`Pronking'}&{`Hop-turn'}&{`Froggy'}\\
				{}&{(Distance)}&{(Yaw angle)}&{(Distance)}\\                
				\hline
				{{Rigid robot}}&{40cm}&{125}&{40}\\
				\hline
				{Compliant robot}&{50cm}&{145}&{50}\\ 
				\hline
				{Incremental ratio}&{ 25$\%$}&{16$\%$}&{25$\%$}\\	           
				\bottomrule  
			\end{tabular}
		\end{table}
		
		\begin{figure}
			\centering
			\includegraphics[width=\columnwidth-0mm]{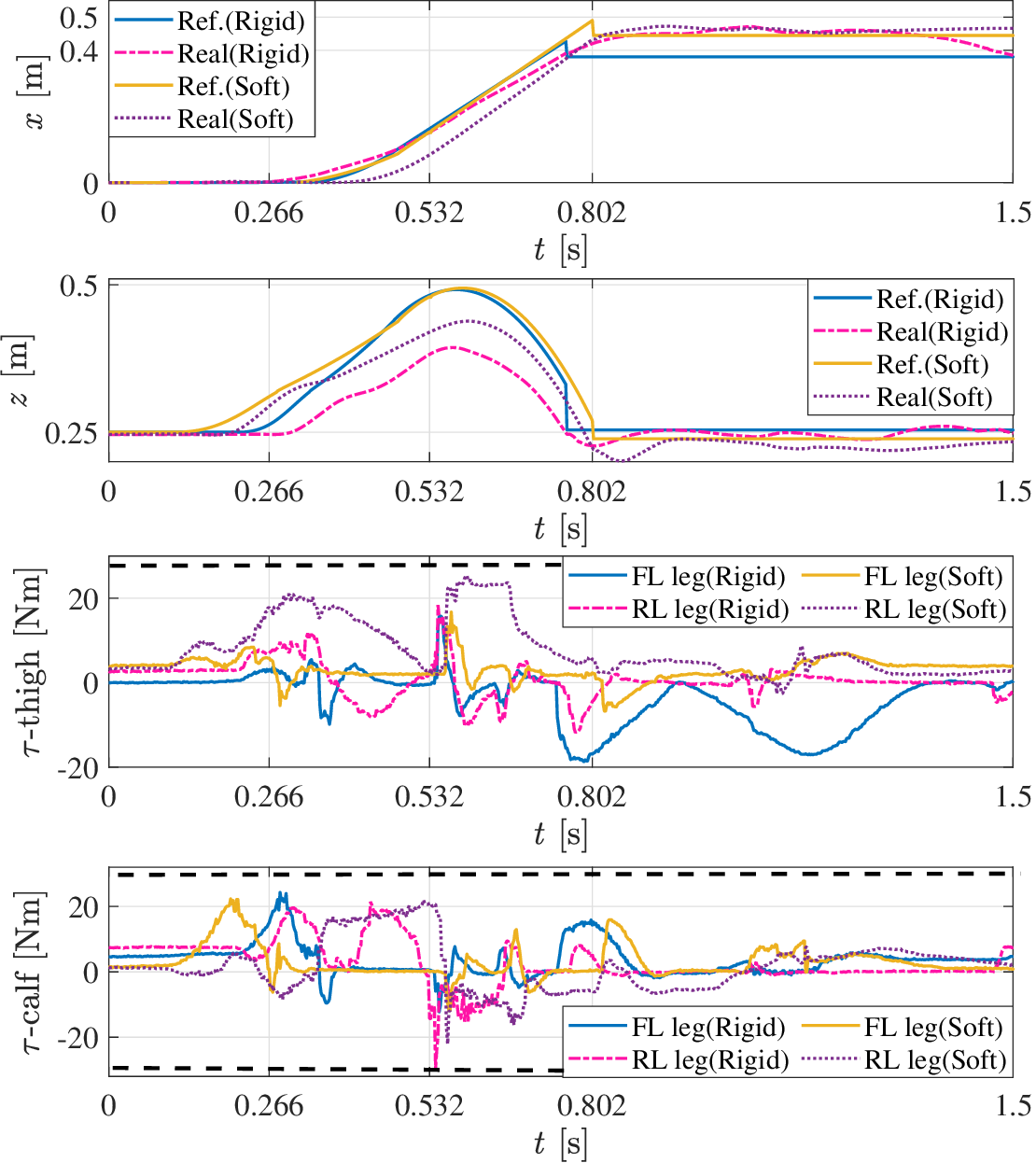}
			\caption{Trajectory profiles for $50\,$cm froggy jumping with the rigid and soft quadrupeds. `0.266', `0.532' and `0.802' separately denote the moments for the half-flight phase with the front legs lifting off, the flight phase with all legs lifting off, and the landing phase with at least two legs touching down when parallel springs are engaged. Dashed lines in the torque profiles mark the torque limits.}
			\label{fig:50cm_froggy_springs_plot}
		\end{figure}
		
		\textit{Remark 4:} When the desired distance of the froggy motion is above $40\,$cm, the calf motor on the rear right leg easily loses control. 
		Thus, we remove the springs on the RR calf joint when performing froggy jumping longer than $40\,$cm. With these settings, the compliant robot can perform froggy jumping up to $50\,$cm, as illustrated in the bottom panel of Fig.~\ref{fig:enhance_tracking_performance}.
		
		The extreme explosive motions with rigid and compliant robots are compared in Table~\ref{table:enhanced_tracking_percentage}. The results demonstrate that our proposed motion generation and control method can exploit parallel compliance to achieve controllable explosive motions with improved performance.

		\subsection{Robustness}
		In this work, instead of using the reference torque generated by the dual-layer TO scheme, we generate the feedforward torque command with the hierarchical control pipeline introduced in Section~\hyperlink{sec_compliance_control}{VI}, endowing robots the robustness against dynamics uncertainties.
		\begin{figure}
			\centering
			\includegraphics[width=\columnwidth-0mm]{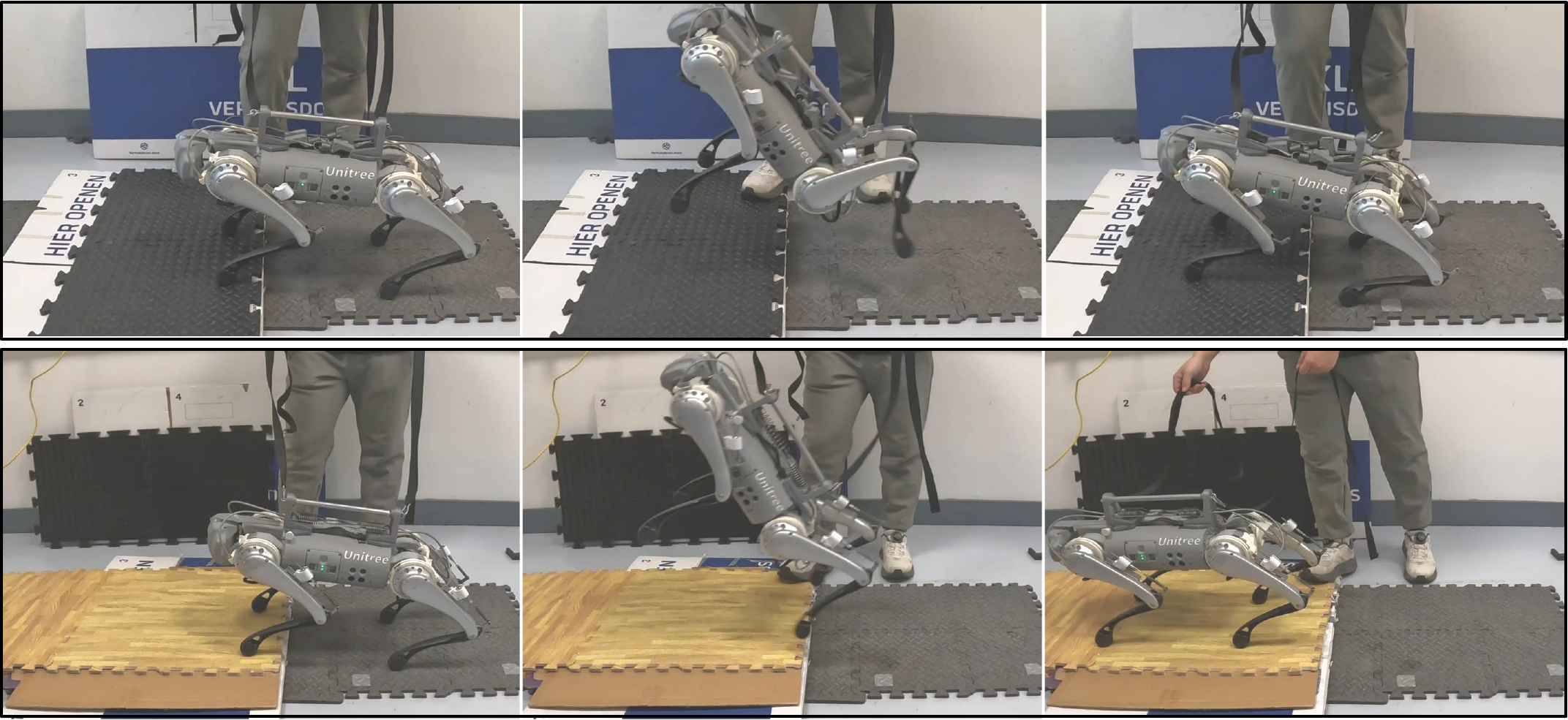}
			\caption{Robust froggy-jumping from the non-coplanar uneven surface with a $6\,$cm height variation. The rigid quadruped (top panel) could not jump onto the platform, whereas the compliant robot  (bottom panel) successfully accomplished the task. In both cases, the robot did not know the terrain information in advance.}
			\label{fig:enhance_robustness_performance}
		\end{figure}

		\subsubsection{Large CoM and angular offset}
		In all the above tests, the robots start moving from flat ground. In this section, we start the robot from an uneven surface with unknown height variation, resulting in a large CoM and angular offset. Taking froggy jumping as an example, the whole-body control scheme enables the rigid and compliant robots to accommodate an uneven surface with a $5\,$cm height variation. Furthermore, the compliant robot with springs engaged can jump onto a $6\,$cm height platform, while the rigid robot without springs engaged can not accomplish this task. 
		The comparison behavior for jumping onto a $6\,$cm high table is visualized in Fig.~\ref{fig:enhance_robustness_performance}, and the movement trajectories are plotted in Fig.~\ref{fig:40cm_froggy_robust}. 
		
		As can be seen from the torque profiles (`$\tau$-calf' plot) in Fig.~\ref{fig:40cm_froggy_robust}, in the stance phase, the rigid robot without springs engaged outputs the largest peak torque to push off the ground, that is, $25\,$Nm, in the FL calf joint. Although the smaller peak torque is found in the soft case, the compliant robot jumped higher, see the real height trajectory after $0.666\,$s (the yellow dotted curve in `\textit{z}' plot) in Fig.~\ref{fig:40cm_froggy_robust}. As a result, the compliant robot landed later and longer than the reference and accomplished the desired jumping task, as illustrated by the second row in Fig.~\ref{fig:enhance_robustness_performance}. 
		
		\begin{figure}
			\centering
			\includegraphics[width=\columnwidth-0mm]{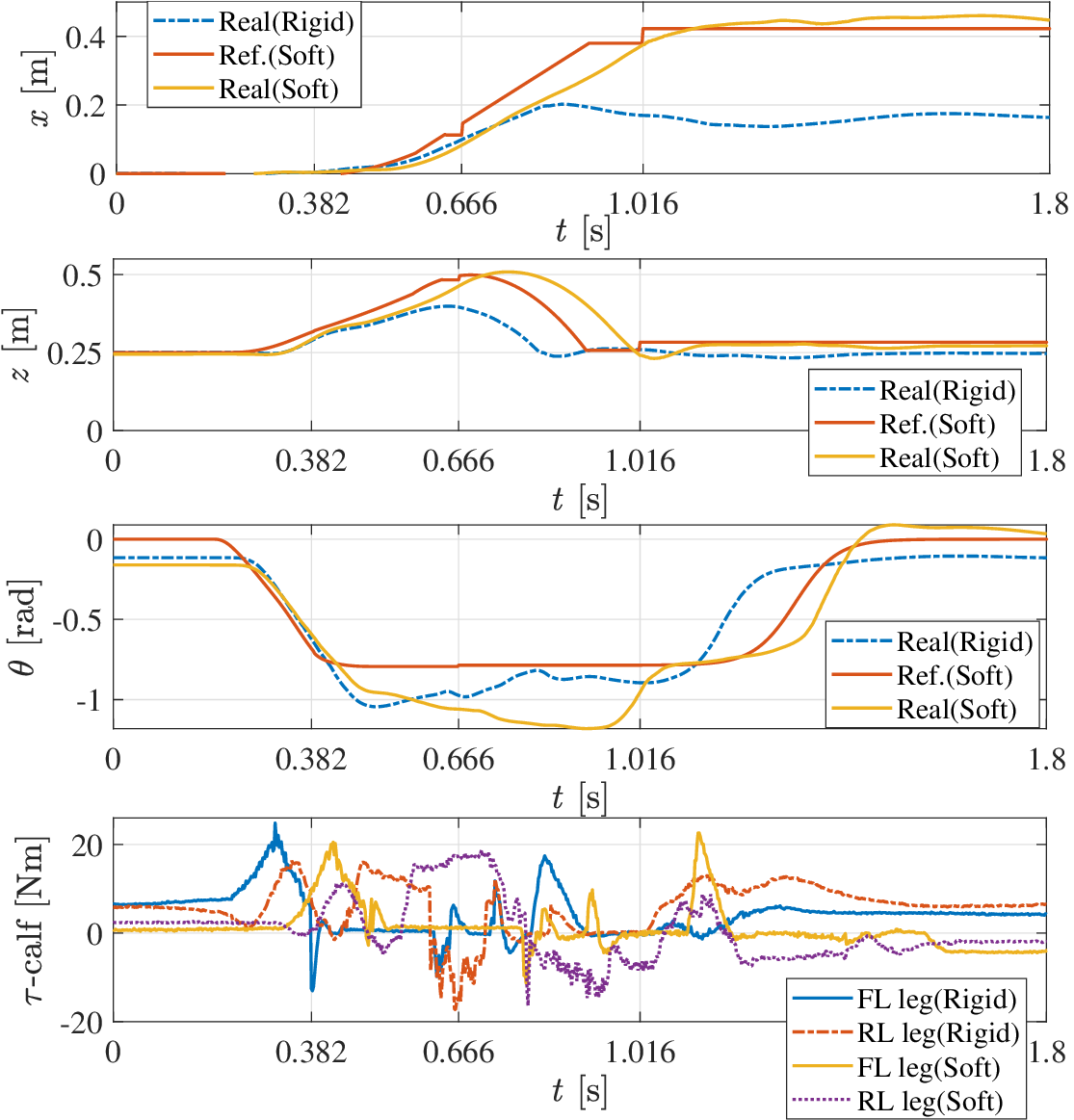}
			\caption{Trajectory profiles for $40\,$cm froggy jumping on the uneven surface. `0.382', `0.666' and `1.016' separately denote the moments for the half-flight phase (the front legs lift off), flight phase (all the legs lift off) and landing phase (at least two legs touch down) when parallel springs are engaged.}
			\label{fig:40cm_froggy_robust}
		\end{figure}

		Extensive experiments reveal that the compliant robot with springs engaged can jump from an uneven surface with a $10\,$cm height variation, resulting in 100$\%$ increase ($10\,$cm \textit{vs} $5\,$cm) than the rigid case. One example of robust froggy jumping is demonstrated at the bottom of Fig.~\ref{fig:snaptshots}.
		
		\subsubsection{Jumping on floating pad}
		The friction status with the ground heavily influences the performance of explosive motion, especially froggy jumping that requires a large body rotation. In this section, we validate the robust jumping from a floating pad. Experiments demonstrate that the compliant robot can achieve a $30\,$cm froggy jumping, as can be seen from the second row of Fig.~\ref{fig:froggy_floating_pad_performance}. It is worth mentioning that although the rigid robot can not jump up to $30\,$cm, it still maintains its balance with the proposed controller.
		
		
		\begin{figure}
			\centering
			\includegraphics[width=\columnwidth-0mm]{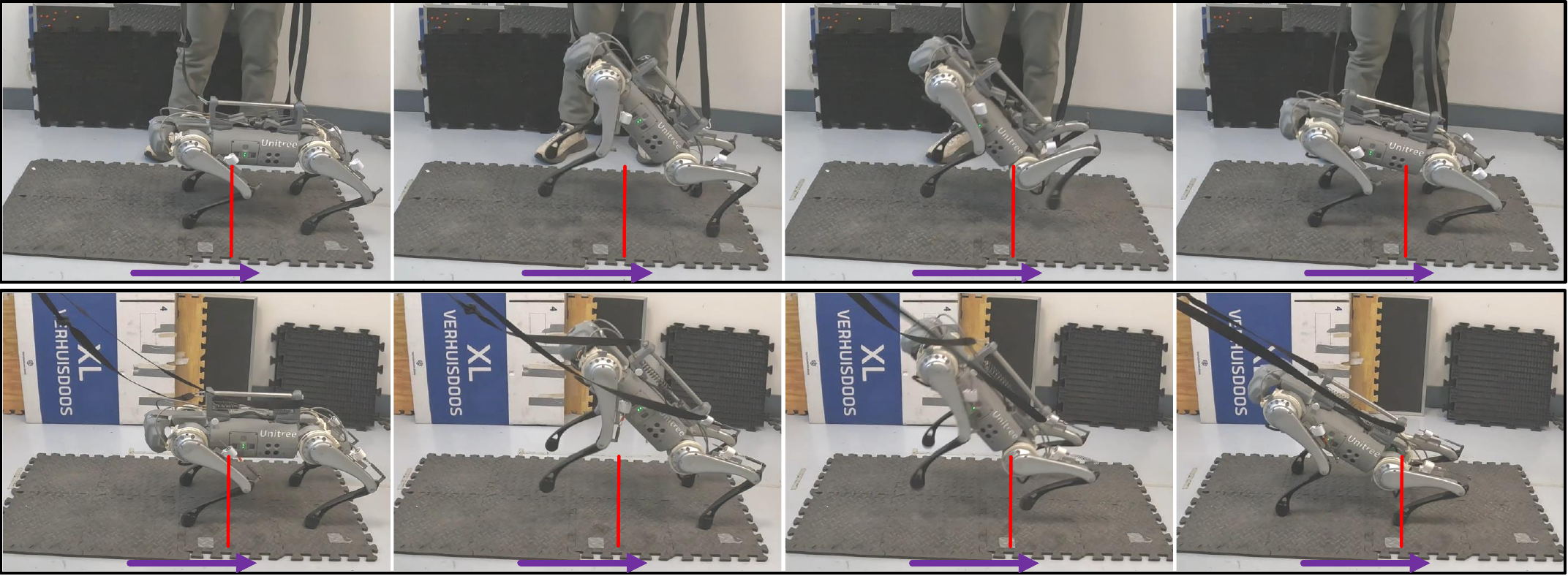}
			\caption{$30\,$cm froggy jumping on a floating pad. The floating pad moves backwards after the robot takes off. The rigid robot (top panel) landed stably but fell behind the desired position. The compliant robot (bottom panel) landed at the desired location. The purple arrows mark the movement direction of the floating pad. The red lines mark the desired landing positions.}
			\label{fig:froggy_floating_pad_performance}
		\end{figure}
		
		\subsubsection{Robustness against modelling error} 
		With the reduced-order template model, the SLIP-based motion planning and the SRB-based MPC result in modelling errors, such as those caused by the ignorance of leg mass.
		However, extensive hardware experiments demonstrate that the robot can accomplish versatile explosive locomotion tasks with decent tracking performance if it does not pursue extreme motion.
		
		In particular, the successful froggy jumping validates the robustness against modelling errors. First, as we mentioned before, the homing height for froggy jumping was reduced to $0.25\,$m. However, the equivalent spring stiffness is computed by assuming the homing height is $0.32\,$m. That is, in the motion planning stage, the modelling error on the parallel compliance is introduced. Secondly, for $50\,$cm froggy jumping, the calf spring in the RR leg is removed, resulting in an asymmetric configuration. Nevertheless, the robot accomplished these tasks successfully. More robust froggy jumping with this asymmetric parallel compliance is attached in the supplementary video. 

		\subsection{Agile dynamic locomotion}
		Although it is not a focus of this work, it is worth mentioning that, with the model-based controller, we can achieve agile locomotion for both the rigid and compliant quadrupedal robots. For example, Fig.~\ref{fig:egov2_agile_loco} displays the contact transition when performing pacing (top), trotting (middle), and trotting-running (bottom) with the compliant quadrupedal robots. Detailed motion can be found in the attached video. 
		\begin{figure}
			\centering
			\includegraphics[width=\columnwidth]{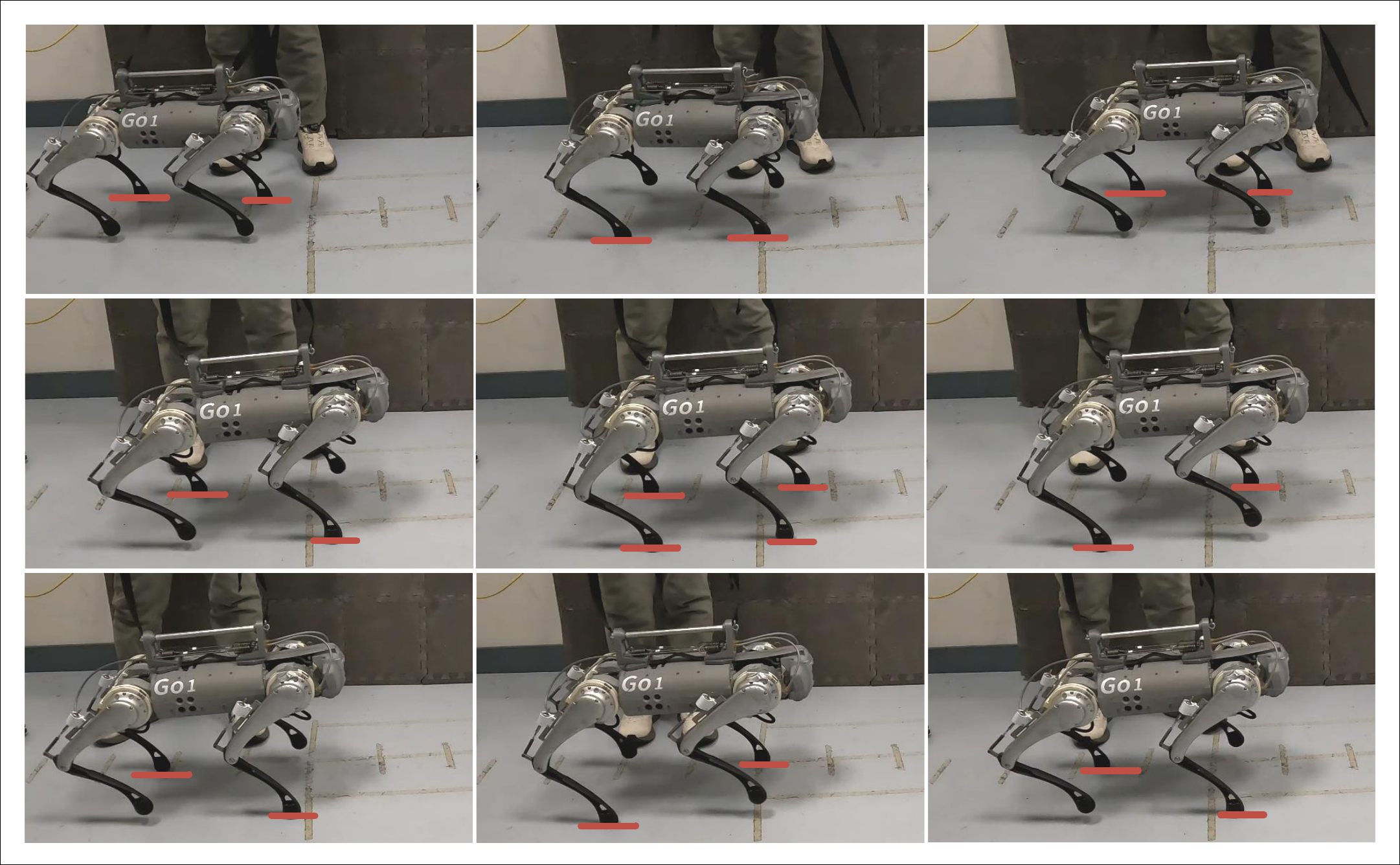}
			\caption{The compliant quadruped with springs engaged performs agile pacing (top panel), trotting (middle panel) and trotting-running (bottom panel). The red lines mark the stance feet. The step periods for pacing, trotting and trotting-running separately are $0.3\,$s, $0.3\,$s and $0.1\,$s, while the step lengths are $3\,$cm, $5\,$cm and $3\,$cm.  }
			\label{fig:egov2_agile_loco}
		\end{figure}


		\section{Discussion and Conclusions} \label{sec_conclusion}
		
		In this work, we realize versatile, robust, and explosive quadrupedal locomotion for both rigid and articulated compliant robots. To this end, we propose a general template model to capture the locomotion dynamics involved with body rotation. By decoupling the motor actuation and parallel compliance, this model is able to exploit the parallel compliance introduced by the mechanical designs. With this reduced-order template model, we propose a dual-layer trajectory optimization strategy to generate multi-modal explosive motion, where quaternion dynamics are incorporated to avoid singularities caused by the large body rotation. Extensive simulation and hardware experiments with rigid and articulated compliant quadrupeds have validated the proposed methodology. The results demonstrate that, by exploiting parallel compliance, enhanced explosive motion with higher robustness can be achieved. 
		
		The following section discusses future work. 
		
		\subsection{Optimal compliance design}
		This work focuses on the modelling and control problem of quadrupedal robots. With this model-based optimization technique, we validate the effectiveness of the parallel compliance, however, without optimizing the compliance design. Although the study in Section~\hyperlink{pronking_performance_spring_setting}{VII-B2} provides a preliminary investigation of the performance under varying rest length and spring stiffness, we do not optimally design the parallel compliance, unlike some pioneering work in mechanical design (\cite{mazumdar2016parallel,sharbafi2019parallel,zhang2024novel}). Since the unidirectional PEA with the constant spring stiffness in joint space is engaged, extra torque is needed to regulate the pose during the flight phase. One example is evidenced by comparing the `FL leg(Soft)' and `FL leg(Rigid)' curves (after 0.532s) in the `$\tau$-thigh' plots from  Fig.~\ref{fig:50cm_froggy_springs_plot}. 
		In future, realizing energy-efficient locomotion with bidirectional, variable stiffness PEA could be a research direction. To this end, the clutch/switchable parallel compliance mechanism (\cite{haufle2012clutched,mathews2022design,liu2018switchable}) could be introduced and the impact-triggered variable stiffness control could be adopted.
		
		Another promising direction is to utilize the up-to-date co-design framework to enhance dynamic locomotion with parallel compliance. Instead of designing the robot body and locomotion controller separately by iteration, co-design engineering (\cite{bravo2024engineering}) enables the co-evolution of the robot body and controller.
		The development of machine learning provides a promising solution. By addressing the sim2real gap, previous work such as (\cite{bjelonic2023learning}) improves the energy efficiency of trotting tasks by optimizing parallel compliance. In the future, it is worth investigating explosive motion with higher energy efficiency via co-design. 
		
		\subsection{Momentum-aware motion planning and control}
		In the current research, we utilized the single-mass model, such as the TD-aSLIP model in motion generation and the single rigid body mode in MPC, to capture the rotation dynamics, ignoring the mass distribution. As a result, the variation of centroidal angular momentum is not considered. In this case, although the quadruped can basically realize the desired movement, the tracking performance is limited, especially for the rigid robot. One example is that a large yaw error occurs when performing the hop-turn task. As demonstrated in the third row of Fig.~\ref{fig:enhance_tracking_performance}, the rigid robot landed with about $70^\circ$ yaw when performing the $90^\circ$ yaw hop-turn.}
	
	To fully capture the locomotion dynamics, more accurate centroidal dynamics or full-body dynamics can be taken into consideration.
	Exiting work, such as
	(\cite{papatheodorou2024momentum,zhou2022momentum,ding2021representation}) has demonstrated that, utilizing the momentum-aware motion planner, the tracking performance of explosive motion with large body rotation, e.g., large yaw movement, could be greatly improved.
	In the future, we will explore the centroidal dynamics or full-body dynamics with parallel compliance in the motion planning stage. In addition, momentum-aware MPC could be incorporated to improve performance further.
	
	\subsection{Agile motion with multi-modal transition}
	In this work, we validate multi-modal explosive motion including pronking, hop-turn, and froggy jumping. At the same time, we also realize agile locomotion, including pacing, trotting and trotting running, with the rigid and articulated compliant quadrupeds. In the future, we are keen to challenge more dynamic locomotion tasks, such as continuous/consecutive jumping (\cite{yang2023continuous,nguyen2022continuous,bellegarda2024quadruped}) or hybrid motion (\cite{li2024cafe,bjelonic2022offline}).
	
	To this end, we can start by building a motion library by leveraging offline trajectory optimization (\cite{bjelonic2022offline,nguyen2022continuous}). Then, based on the real-time state estimation, we select/synthesize the desired task, without requiring online trajectory optimization. To move further, we can train a goal-conditioned neural network, learning the torque command policy directly and fully releasing the computational burden to solve the optimization problem. In particular, we would like to exploit parallel compliance when achieving these challenging tasks.

	\bibliographystyle{SageH}
	\bibliography{sty/mainsage.bib}

	
	
	
	
\end{document}